\documentclass[conference]{IEEEtran}
\IEEEoverridecommandlockouts
\usepackage{cite}
\usepackage{amsmath,amssymb,amsfonts}
\usepackage{algorithmic}
\usepackage{graphicx}
\usepackage{textcomp}
\usepackage{array}
\usepackage[table,xcdraw]{xcolor}
\usepackage{lipsum}

\def\BibTeX{{\rm B\kern-.05em{\sc i\kern-.025em b}\kern-.08em
    T\kern-.1667em\lower.7ex\hbox{E}\kern-.125emX}}
\begin{document}

\title{Tracking Noisy Targets: A Review of Recent Object Tracking Approaches\\
}

\author{\IEEEauthorblockN{Mustansar Fiaz\IEEEauthorrefmark{1},
		Arif Mahmood\IEEEauthorrefmark{2} and
		Soon Ki Jung\IEEEauthorrefmark{3}}
	\IEEEauthorblockA{\IEEEauthorrefmark{1}\IEEEauthorrefmark{3}School of Computer Science and Engineering, \\
		Kyungpook National University, Republic of Korea\\
		\IEEEauthorrefmark{2}Department of Computer Science and Engineering, \\
		Qatar University, Qatar\\
		Email: \IEEEauthorrefmark{1}mustansar@vr.knu.ac.kr,
		\IEEEauthorrefmark{2}rfmahmood@gmail.com,
		\IEEEauthorrefmark{3}skjung@knu.ac.kr}}

\maketitle

\begin{abstract}
Visual object tracking is an important computer vision problem with numerous real-world applications including human-computer interaction, autonomous vehicles, robotics, motion-based recognition, video indexing, surveillance and security. In this paper, we aim to extensively review the latest trends and advances in the tracking algorithms and evaluate the robustness of trackers in the presence of noise. The first part of this work comprises a comprehensive survey of recently proposed tracking algorithms. We broadly categorize trackers into correlation filter based trackers and the others as non-correlation filter trackers. Each category is further classified into various types of trackers based on the architecture of the tracking mechanism. In the second part of this work, we experimentally evaluate tracking algorithms for robustness in the presence of additive white Gaussian noise. Multiple levels of additive noise are added to the Object Tracking Benchmark (OTB) 2015, and the precision and success rates of the tracking algorithms are evaluated. Some algorithms suffered more performance degradation than others, which brings to light a previously unexplored aspect of the tracking algorithms. The relative rank of the algorithms based on their performance on benchmark datasets may change in the presence of noise. Our study concludes that no single tracker is able to achieve the same efficiency in the presence of noise as under noise-free conditions; thus, there is a need to include a parameter for robustness to noise when evaluating newly proposed tracking algorithms.
\end{abstract}

\begin{IEEEkeywords}
Visual Object Tracking; Robustness of Tracking Algorithms, Surveillance, Security,
Tracking Evaluation
\end{IEEEkeywords}
\section{Introduction}
Visual Object Tracking (VOT) is a promising but difficult computer vision problem. It has attained much attention due to its widespread use in applications such as autonomous vehicles \cite{brown2017safe,laurense2017path}, traffic flow monitoring \cite{ tian2011video}, surveillance and security \cite{sivanantham2016object}, robotics \cite{onate2017tracking}, human machine interaction \cite{severson2017human}, medical diagnostic systems \cite{walker2017systems} and activity recognition \cite{aggarwal2014human}. 
VOT has remained an active research topic due to both the opportunities and the challenges. Remarkable efforts have been made by research community in the past few decades, but VOT has much potential still to be explored. The difficulty of VOT lies in the myriad challenges, such as occlusion, clutter, variation illumination, scale variations, low resolution targets, target deformation, target re-identification, fast motion, motion blur, in-plane and out-of-plane rotations, and target tracking in the presence of noise \cite{wu2013online, wu2015object}.\\
Typically, object tracking is the process of identifying a region of interest in a sequence of frames. Generally, the object tracking process is composed of four modules, including target initialization, appearance modeling, motion estimation and target positioning. Target initialization is the process of annotating object position, or region of interest, with any of the following representations: object bounding box, ellipse, centroid, object skeleton,  object contour, or object silhouette. Usually, an object bounding box is provided in the first frame of a sequence and the tracker is to estimate target position in the remaining frames in the sequence. Appearance modelling is composed of identifying visual object features for better representation of a region of interest and effective construction of mathematical models to detect objects using learning techniques. Motion estimation is the process of estimating the target location in subsequent frames. The target positioning operation involves maximum posterior prediction, or greedy search. Tracking problems can be simplified by constraints imposed on the appearance and motion model. A large variety of object trackers have been proposed to answer questions about what to track, whether there is suitable representation of the target for robust tracking, what kind of learning mechanisms are appropriate for robust tracking, and how appearance and motion can be modeled.\\
Despite the fact that much research has been performed on object tracking, no up-to-date survey exists to provide a comprehensive overview that might give researchers insight about recent trends and advances in the field.
Yilmaz et al. \cite{yilmaz2006object} provided an excellent overview of tracking algorithms, feature representations and challenges. However, the field has greatly advanced in recent years.  Cannons et al. \cite{cannons2008review} covered the fundamentals of object tracking problems, and discussed the building blocks for object tracking algorithms, the evolution of feature representations and different tracking evaluation techniques. Smeulders et al. \cite{smeulders2014visual} compared the performance of tracking algorithms. Li et al. \cite{li2013survey} and Yang et al. \cite{yang2011recent} discussed object appearance representations, and performed surveys for online generative and discriminative learning. Most of the surveys are somewhat outdated and subject to traditional tracking methods. The performance of tracking algorithm was boosted by the inclusion of deep learning techniques and none of the existing surveys cover recent trackers.\\
The objective of the current study is to provide an overview of the recent progress and research trends and to categorize existing tracking algorithms. Our motivation is to provide  interested readers an organized reference about the diverse  tracking algorithms being developed, and to help them find research gaps and provide insights for developing new tracking algorithms. Our study also enables a reader to select appropriate trackers for specific applications, especially for real-world scenarios that involve visual noise.\\
Numerous tracking algorithms have been proposed to handle different object tracking challenges. For example Zhang et al. \cite{zhang2014partial}, Pan and Hu \cite{pan2007robust} and Yilmaz et al. \cite{yilmaz2004contour} proposed tracking algorithms to handle occlusion in videos.  Similarly, several tracking algorithms have been developed to tackle illumination variations such as  those by Zhang et al. \cite{zhong2012robust}, Adam et al. \cite{adam2006robust} and  Babenko et al. \cite{babenko2009visual}. Moreover, Mei et al. \cite{mei2009robust}, Kalal et al. \cite{kalal2010pn}, and Kwon et al. \cite{kwon2010visual} proposed trackers to handle the problem of cluttered backgrounds.  Thus, various tracking techniques have been developed to deal with different tracking challenges, however the robustness of trackers to noise has not been thoroughly evaluated. Though the benchmarks may contain some noisy sequences, robustness to noise has not been thoroughly tested. Thus, there is need to test trackers in the presence of synthetic noise. In this work we perform a comprehensive evaluation of tracking algorithms on white Gaussian noise added to OTB2015.\\
Digital noise appears as a  grainy effect or speckled colour in images. Noise is unavoidable and undesirable byproduct of the image acquisition process. Noise may get added due to an image for several reasons, such as over-exposure, poor focus, the presence of magnetic field generated by electronic circuits, the dispersion of light by a lens, light intensity variations, and  object blur due to camera or object motion. There can be many other types of noise caused by the environment,  for example,  fog, rain, shadows, and bright spots. Noise can negatively effect the performances of visual object trackers. Therefore, evaluating the robustness of trackers to different types of noise is essential. This evaluation will give a better  understanding of the impact of different noise types on different trackers, and will provide insight for selecting suitable trackers for a given scenario.  We explore this new research direction, and produce a benchmark where sequences include more rigorous noise.  Ideally, a tracker must be able to handle various types of commonly-occurring noise to perform robust object tracking. In this work, we evaluate the robustness of the most recent tracking algorithms to additive Gaussian noise. To the best of our knowledge, tracking noisy targets has not been addressed before us.\\
The rest of the paper is organized as follows: Section II describes  related work; the  classification of recent tracking algorithms is explained in section III with the brief introduction of the selected state-of-the-art trackers; in section IV experiments and evaluation are performed on various levels of noise in OTB2015; and in  Section V the conclusion and future directions are discussed.
\section{Related Work}
The research community has shown  keen interest in Visual Object Tracking (VOT), and has developed a number of state-of-the-art tracking algorithms. Therefore, a review of research methodologies and techniques will be helpful in organizing domain knowledge. Visual object tracking algorithms can be categorized as single-object vs. multiple-object trackers, generative vs. discriminative, context-aware vs. non-aware, and   online vs. offline learning algorithms. Single object trackers \cite{leang2018line, tahir2016single, lee2016globally} are the algorithms tracking only one object in the sequence, while multi-object trackers \cite{yang2005fast, berclaz2011multiple, perera2006multi, leal2017tracking} simultaneously track multiple targets and follow their trajectories.  In generative models, the tracking task is carried out via searching the best-matched window, while discriminative models discriminate target patch from the background  \cite{qin2014object, yu2008online, yang2011recent}. In the current paper, recent tracking algorithms are classified as Correlation-Filter based Trackers (CFTs) and Non-Correlation Filter based Trackers (NCFTs). It is obvious from the names that CFTs \cite{gundogdu2016evaluation, sui2016real, chen2017visual}  utilize correlation filters, and non-correlation  trackers use other techniques \cite{gu2010efficient, khan2011robust, gong2011multi}.\\
Yilmaz et al. \cite{yilmaz2006object} presented a taxonomy of tracking algorithms and discussed tracking methodologies, feature representations, data association, and various challenges. Yang et al. \cite{yang2011recent} presented an overview of the local and global feature descriptors used to present object appearance, and reviewed  online learning techniques such as generative versus discriminative,  Monte Carlo sampling techniques, and integration of contextual information for tracking. Cannons \cite{cannons2008review}  discussed  object tracking components initialization, representations, adaption, association and estimation. He discussed the advantages and disadvantages of different feature representations and their combinations.  Smeulders et al. \cite{smeulders2014visual} performed  analysis and evaluation of different trackers with respect to a variety of tracking challenges.  They found sparse and local features more suited to handle illumination variations, clutter, and occlusion. They used various evaluation techniques,  such as survival curves, Grubs testing, and Kaplan Meier statistics, and provided evidence that  F-score is the best  measure of tracking performance. Li et al. \cite{li2013survey} gave a detailed summary of target appearance models. Their study included local and global feature representations, discriminative, and generative, and hybrid learning techniques.\\
Some relatively limited or focused reviews include the following works.
Qi et al. \cite{liu2014survey} focused on classification of online single target trackers. Zhang et al. \cite{zhang2013sparse} discussed tracking based on sparse coding, and classified sparse trackers. Ali et al. \cite{ali2016visual} discussed some  classical  tracking algorithms. Yang et al. \cite{yang2009context}  considered  context of tracking scene considering auxiliary objects \cite{yang2006intelligent} as the target context. Chen et al. \cite{chen2015experimental} examined only  CFTs. 
Arulampalam et al. \cite{arulampalam2002tutorial} presented  Bayesian tracking methods using particle filters. Most of these studies are outdated or consider only few algorithms and thus are limited in scope. In contrast, we present a more comprehensive survey of recent contributions. We classify tracking algorithms as CFTs and NCFTs. Furthermore, we evaluate  state-of-the-art trackers in the presence of noise to test their robustness when tracking noisy targets.
\section{Classifications of Tracking Algorithms}
In this section, we study recent tracking algorithms, most of them were proposed during the last three years. Each algorithm presents a different method to exploit target structure for predicting target location in a sequence. By analyzing the tracking procedure, we arrange these algorithms in a hierarchy and classify them into  different categories. We  classify the trackers into two main categories: Correlation Filter Trackers (CFT) and Non-correlation Filter Tracker (NCFT) also referred as traditional trackers, with a number of subcategories in each class. 
\begin{figure*}[]
\centering
\includegraphics[scale=0.4]{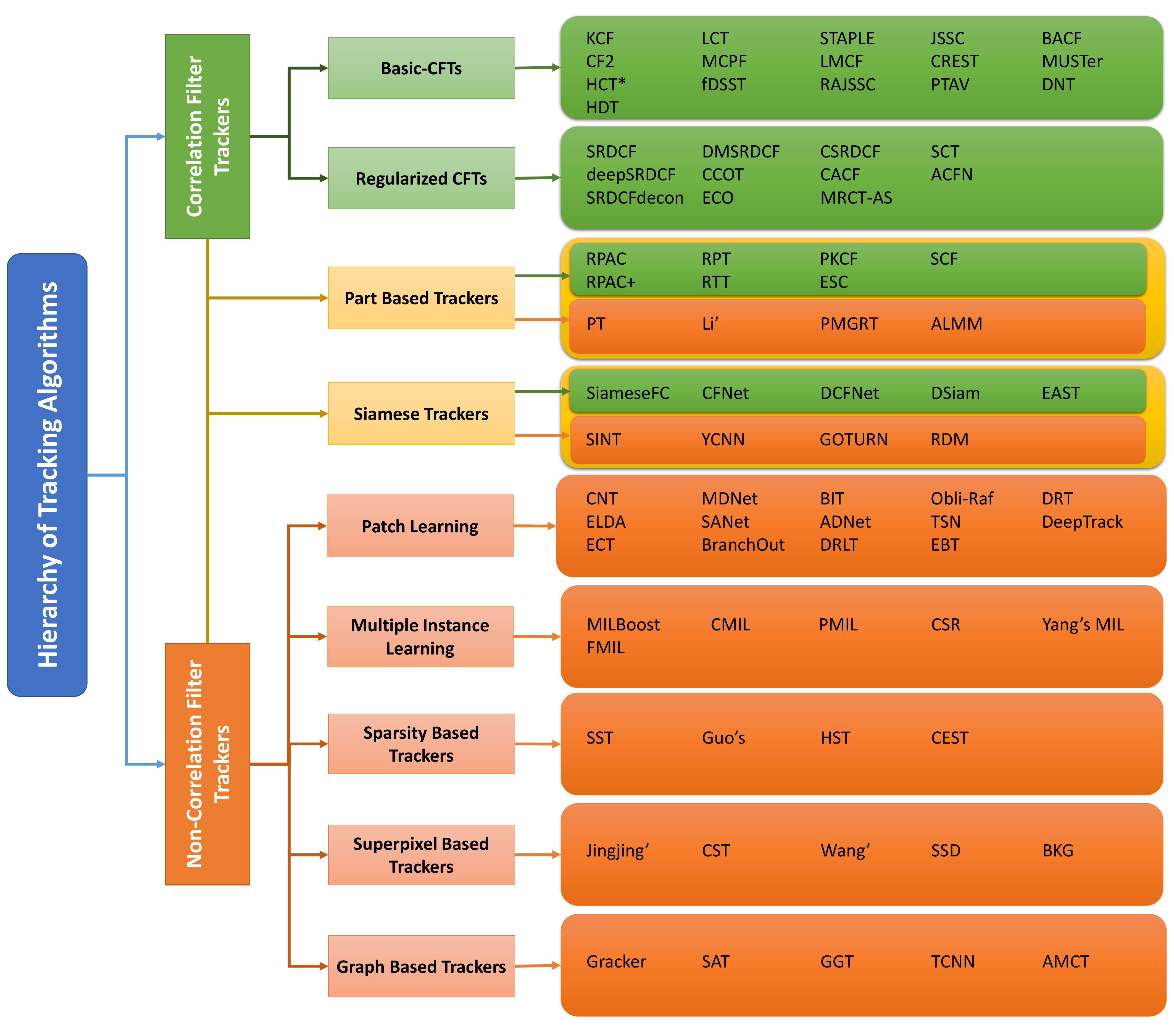}
\caption{Taxonomy of tracking algorithms}
\label{Hier1}
\end{figure*}
\subsection{Correlation Filter Trackers}
Correlation filters (CF) have been actively used in various computer vision applications such as object recognition \cite{felzenszwalb2010object}, image registration \cite{essannouni2006fast}, face verification \cite{savvides2002face}, and action recognition \cite{rodriguez2008action}. In object tracking, CF have  been used to improve robustness and efficiency.  Initially, the requirement of training made CF inappropriate for online tracking \cite{bolme2010visual}.  In the later years, development  of Minimum Output of Sum of Squared Error (MOSSE) filter \cite{bolme2010visual}, which allows for  efficient adaptive training, changed the situation. MOSSE is an improved version of  Average Synthetic Exact Filter (ASEF) \cite{bolme2009average}. Later on,  many state-of-the-art CFT based on MOSSE were proposed. Traditionally, the aim of designing inference of CF is to yield  response map that has low value for background and high values for region of interest in the scene. One such tracker is Circulant Structure with Kernal (CSK) tracker \cite{henriques2012exploiting}, which exploits circulant structure of the target appearance and  is trained using kernel regularized least squares method.\\
\begin{figure*}[]
\centering
\includegraphics[scale=0.5]{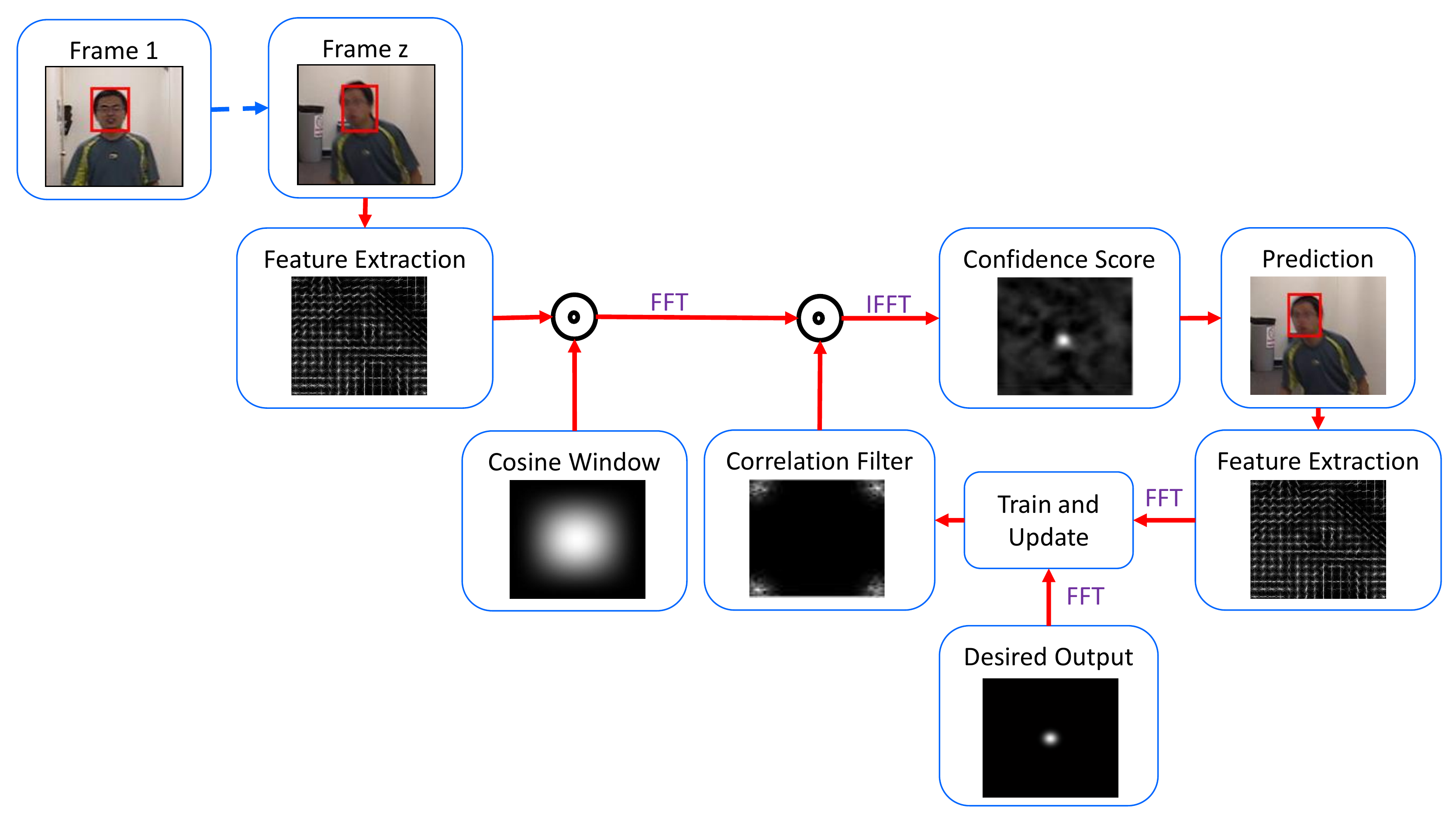}
\caption{Correlation Filter Tracking Framework \cite{chen2015experimental}}
\label{CF1}
\end{figure*}
CF-based tracking schemes perform computation in the frequency domain to manage computational cost. Figure \ref{CF1} shows the general framework of these algorithms. Correlation filters are initialized  with a target patch cropped from the target location in the initial frame of the sequence. During tracking, a patch containing the target location is estimated in the current frame based on the target location in the previous frame. To effectively represent target appearance, an appropriate type of features may be extracted from the selected patch. Boundaries are smoothed by applying a cosine filter. The response map is computed using element-wise multiplication of the adaptive learning filter and the estimated target patch,  and by using a Fast Fourier Transform (FFT) in the frequency domain. The Inverse FFT (IFFT) is applied over the response map to obtain confidence map in the spatial domain. New target position is estimated at the maximum confidence score. At the outcome, the target appearance at the newly predicted location is updated by extracting features and updating correlation filters.\\
Let $h$ be a correlation filter and $x$ be the estimated patch in the current frame, which may consist of the extracted features or the raw image pixels. By the convolution theorem, element-wise multiplication in the frequency domain is the same as convolution in spatial domain.
\begin{equation} \label{eq:cf1}
x \otimes h= \mathfrak{F}^{-1}(\widehat {x} \odot \widehat {h}^\ast),
\end{equation}
in the above equation, where $\otimes$ represents convolution, $\mathfrak{F}^{-1}$ denotes the IFFT, $\odot$ means element-wise multiplication and $\ast$ is the complex conjugate. Equation \ref{eq:cf1} yields a confidence map between $x$ and $h$. To update the correlation filter, the estimated target around the maximum confidence position is selected. Assume $y$ is the desired output. Correlation filter $h$ must satisfy for new target appearance $z$ as:
\begin{equation} \label{eq:cf2}
y= \mathfrak{F}^{-1}(\widehat z \odot \widehat h\ast),
\end{equation}
hence
\begin{equation} \label{eq:cf3}
\widehat h\ast =\frac{\widehat y}{\widehat z},
\end{equation}
where $\widehat y$ denotes the desired output $y$ in frequency domain and division operation is performed during element-wise multiplication. FFT reduces the computational cost, as circulant convolution has a complexity of $O(n^4)$ for image size $nxn$ while FFT require only $O(n^2 \log n)$.\\
CF-based tracking frameworks face different difficulties, such as the training of the target appearance, as it may change over time. Another challenge is the selection of  an efficient feature representation for CFTs, as powerful features may improve the performance of CFTs.  Another important challenge for CFTs is scale adaption, as the size of correlation filters are fixed during tracking. A target may change its scale over time. Furthermore, if the target is lost then it cannot be recovered again. CF-based trackers are further divided into the categories  \textit{k}-CF, regularized operator, part based, and Siamese CFTs, as explained below.\\
\subsubsection{Basic Correlation Filter based Trackers}
Basic-Correlation Filter based Trackers (B-CFTs) are  tackers that use high-speed tracking with Kernelized Correlation Filters (KCF) \cite{henriques2015high} as their baseline tracker. Trackers may use different features such as the HOG \cite{mcconnell1986method}, colour names (CN) \cite{danelljan2014adaptive} and deep features using Convolutional Neural Networks (CNN) \cite{simonyan2014very}, Recurrent Neural Networks (RNN) \cite{williams1989learning} and residual features \cite{he2016deep}. Some of the B-CFTs also perform scale estimation of target using pyramid strategies \cite{danelljan2017discriminative}.\\
A KCF \cite{henriques2015high} algorithm performs tracking using Gaussian kernel function for distinguishing between target object and its surroundings. A KCF tracker uses HOG descriptors with a cell size of 4. During tracking, an image patch is extracted greater then the size of the target  estimated in the previous frame. HOG features are computed for that patch and a response map is computed by applying learned correlation filter on input features in Fourier domain. A new position of the target at the position of maximum confidence score in the confidence map obtained is predicted by applying inverse Fourier transform on response map. A new patch containing object  is then cropped and HOG features are recomputed to update the CF. \\
Ma et al.\cite{ma2015hierarchical} exploited  rich hierarchical Convolutional Features in Correlation Filter (CF2) for visual tracking. For every subsequent frame, CF2 crops the search region centered at the target estimated in the previous frame. Three hierarchical convolutional features are extracted using VGG-Net \cite{simonyan2014very} which is trained on ImageNet \cite{deng2009imagenet} to exploit target appearance. 
An independent adaptive correlation filter is applied for each CNN feature, and response maps are computed. A coarse to fine translation estimation strategy is applied over the set of correlation response maps to estimate the new target position. Adaptive hierarchical correlation filters are updated on newly-predicted target location. Ma et al. \cite{ma2017robust} also proposed Hierarchical Correlation Feature based Tracker (HCFT*), which is an extension of CF2 that integrates re-detection and scale estimation of target.\\
The Hedged Deep Tracking  (HDT) \cite{qi2016hedged} algorithm takes advantage of multiple levels of CNN features. In HDT, authors hedged many weak trackers  together to attain a single strong tracker. During tracking, the target position at the previous frame is utilized to crop a new image 
to compute six deep features using VGGNet. 
Deep features were exploited to individual CF to compute response maps also known as weak experts. The target position is estimated by each weak tracker, and the loss for each expert is also computed.  A standard hedge algorithm is used to estimate the final position. All weak trackers are hedged together into a strong single tracker by applying an adaptive online decision algorithm. Weights for each weak tracker are updated during online tracking. In an adaptive Hedge algorithm, a regret measure is computed for all weak trackers as a weighted average loss. A stability measure is computed for each expert based on the regret measure. The hedge algorithm strives  to minimize the cumulative regret of weak trackers depending upon its historical information. 
The Long-term Correlation Tracking (LCT) \cite{ma2015long} algorithm involves exclusive translation and scale estimation of the target using correlation filters and online re-detection of the target during tracking by using a random fern classifier \cite{ozuysal2007fast}. In LCT algorithms, the search window is cropped on the previously estimated target location and a feature map is computed. Translation estimation is performed using adaptive translation correlation filters. A target pyramid is generated around the newly estimated translation location of target, and scale estimation is done using a separate regression correlation model. The LCT tracking algorithm performs re-detection in the case of failure. If the  estimated target score is less then a threshold,  re-detection is then performed using online random fern classifier \cite{ozuysal2007fast}. Average response is computed using posteriors from all the ferns. LCT selects the positive samples to predict new patch as target by using the \textit{k}-nearest neighbor (KNN) classifier.\\
The Multi-task Correlation Particle Filter (MCPF) \cite{Zhang_2017_CVPR} is based on a particle filter framework. 
The MCPF shepherd particles in the search region representing all the circulant shifts which covers all the states of target object. The MCPF computes response map particles, and target position is estimated as weighted sum of the response maps\\
Discriminative Scale Space Tracking (DSST) \cite{danelljan2017discriminative} learns independent correlation filters for precise translation and scale estimation.  Scale estimation is done by learning the target sample at various scale variations. In proposed framework, the target translation is first estimated by applying a standard translation filter to every incoming frame. After translation estimation, the target size is approximated by employing trained scale filter at the target location obtained by the translation filter. 
This way, the proposed strategy learns the target appearance induced by scale rather than by using exhaustive target size search methodologies. 
The author further improved the computational performance and target search area in fast DSST (fDSST) without sacrificing  the accuracy and robustness of the tracker by using sub-grid interpolation of correlation scores.\\
The Sum of Template And Pixel-wise LEarners (STAPLE) \cite{bertinetto2016staple} algorithm exploits the inherent structure of each patch representation by maintaining two separate regression problems.  The tracking design takes advantage of two complementary factors from two different patch illustrations to train a model. HOG features  and global color histograms are used to represent the target. In the colour template, foreground and background regions are computed at previously estimated location. The frequency of each colour bin for object and background are updated, and a regression  model for colour template is computed. A per-pixel score is calculated in the search area  based on the previously estimated location, and the integral image is used to compute response, while for the HOG template, HOG features are extracted at the  position predicted in the previous frame, and CF are updated. At every incoming frame, a search region centered at previous predicted location is extracted, and their HOG features are convolved with CF to obtain a dense template response. Target position is estimated by a linear combination of both template and histogram response scores. Final estimated location is influenced by the model which has more scores.  Wang et al. \cite{wang2017large} proposed Large Margin object tracking with Circulant Features (LMCF) which increases the discriminative ability and introduces multimodel target detection to avoid drift.\\
Joint scale-spatial correlation tracking with adaptive rotation estimation (RAJSSC) \cite{zhang-ras2015joint} represents target appearance via spatial displacements, scale changes, and rotation transformations. JSSC \cite{zhang-JSSC2015robust} performs exhaustive search for scale and spatial estimation via block circulant matrix. For rotation orientation, the target template is transferred to the Log-Polar coordinate system and uniform CF framework is used for rotation estimation. 
The Convolutional RESidual learning for visual Tracking (CREST) algorithm \cite{song-iccv17-CREST} utilizes residual learning \cite{he2016deep} to adapt target appearance and also performs scale estimation by searching patches at different scales. During  tracking, the search patch is cropped at previous location, and convolutional features are computed. Residual and base mapping are used to compute the response map. The maximum response value gives the newly estimated target position. Scale is estimated by exploring different scale patches at newly estimated target center position.\\
The Parallel Tracking And Verifying (PTAV) \cite{fan2017parallel} framework consists of two major components, i.e. tracker and verifier. Tracker module is responsible for computing the real time inference and estimate tracking results, while the verifier is responsible for checking whether the results are correct or not. Kiani et al. \cite{kiani2017learning} exploited the background patches and proposed Background Aware Correlation Filter (BACF) tracker. Wang et al. \cite{wang2016stct} proposed a framework to fine tune best online tracker from sequential CNN learners via sampling in such a way that correlation among learned deep features is not high. The Multi-Store tracker (MUSTer) \cite{hong2015multi} is based on the Atkinson-Shiffrin Memory Model (ASMM) \cite{atkinson1968human}, comprising of short term store and long term store to aggregate image information and perform tracking. Short term storage involves an Integrated Correlation Filter (ICF) to incorporate spatiotemporal consistency, while long term storage involves integrated RANSAC estimation and key point match tracking to control the output. A Dual deep network (DNT) \cite{chi2017dual} is based on Independent Component Analysis with Reference (ICA-R) \cite{lu2006ica}. DNT exploits high-level features and lower-level features to encode semantic and appearance context. \\
\subsubsection{Regularized Correlation Filter Trackers}
Discriminative Correlation Filter (DCF)-based trackers are limited in their detection range because they require filter size and patch size to be equal. The DCF may learn the  background for irregularly-shaped target objects. The DCF is formulated from periodic assumption, learns from a set of training samples, and  thus may learn negative training patches. DCF response maps have accurate scores close to the centre, while other  scores are influenced due to periodic assumption, thus degrading DCF performance. Another limitation of DCFs is that they are restricted to only a fixed search region. DCF trackers performed poorly on a target deformation problem due to over fitting of model due caused by learning from target training samples but missing the negative samples. Thus, the tracker fails to re-detect in case of occlusion.  A larger search region may solve the occlusion problem but the model will learn background information which  degrades the  discrimination power of the tracker. Therefore, there is a need to incorporate a measure of regularization for these DCF limitations.\\
Danelljan et al.\cite{danelljan2015learning} presented Spatially Regularized DCF (SRDCF) by introducing spatial regularization in DCF learning. During  tracking, a regularization component weakens the background information. 
Spatial regularization measures the weights of filter coefficients based on spatial information. The background is suppressed by assigning higher values to the filter coefficients that are located outside of the target territory and vice versa. 
The SRDCF framework has been updated by using deep CNN features in deepSRDCF \cite{danelljan2015convolutional}. The SRDCF framework has also been modified to handle contaminated training samples in SRDCFdecon \cite{danelljan2016adaptive}. It down weights corrupted training samples and estimate good quality samples. SRDCFdecon extracts training samples from previous frames and then assign higher weights to correct training patches. The appearance model and the training sample weights are learned jointly in SRDCFdecon.\\
Recently, deep motion features have been used for activity recognition \cite{gkioxari2015finding, kiani2015correlation}. Motion features are obtained from  information obtained directly  from optical flow applied to images. A CNN is then applied to optical flow to get deep motion features. 
Gladh et al.\cite{gladh2016deep} presented Deep Motion SRDCF (DMSRDCF) algorithm which fused deep motion features along with hand-crafted appearance features using SRDCF as baseline tracker. Motion features are computed as reported by \cite{cheron2015p}. Optical flow is calculated on each frame  on previous frame using an algorithm by Brox \cite{brox2004high}. 
The x component, y component and magnitude of optical flow constitute three channels in the flow map, which is normalized between 0 and 255 and fed to the CNN to compute deep motion features.\\
Danelljan et al.\cite{DanelljanECCV2016} proposed  learning multi-resolution feature maps, which they name as Continuous Convolutional Operators for Tracking (CCOT). The convolutional filters are learned in a continuous sequence of resolutions which generates a sequence of response maps. These multiple response maps are then fused to obtain final unified response map to estimate target position.\\ 
The Efficient Convolution Operators (ECO) \cite{DanelljanCVPR2017} tracking scheme is an improved version of CCOT. The CCOT learns a large number of filters to capture target representation from high dimensional features, and updates the filter for every frame, which involves training on a large number of sample sets. 
In contrast, ECO constructs a smaller set of filters to efficiently capture target representation using matrix factorization. 
The CCOT learns over consecutive samples in a sequence which forgets target appearance over a long period of time thus causes overfitting to the most recent appearances and   leading to high computational cost. In contrast, ECO uses a Gaussian Mixture Model (GMM) to represent diverse target appearances. 
Whenever a new appearance is found during tracking, a new GMM component is initialized. Declercq and Piater  online algorithm \cite{declercq2008online} is used to update GMM components. If the maximum number of components exceeds a limit, then a GMM component with minimum weight is discarded if its weight is less than a threshold value. Otherwise, the two closest components are merged into one component.\\ 
The Channel Spatial Reliability for DCF (CSRDCF) \cite{Lukezic_CVPR_2017} tracking algorithm integrates channel and spatial reliability with DCF. 
Training patches also contain non-required background information in addition to the required target information. Therefore, DCFs may also learn background information, which may lead to the drift problem.
In CSRDCF, spatial reliability is ensured by estimating a spatial binary map at current target position to learn only target information. 
Foreground and background models retained as colour histogram are used to compute appearance likelihood using Bayes' rule. 
A constrained CF is obtained by convolving the CF with spatial reliability map that indicates which pixels should be ignored. 
Channel reliability is measured as a product of channel reliability measure and detection reliability measures. The channel reliability measure is the maximum response of channel filter. Channel detection reliability in response map is computed from the ratio between the second and first major modes, clamped at 0.5. Target position is estimated at maximum response of search patch features and the constrained CF, and is weighted by channel reliability.\\ 
Mueller et al. \cite{mueller2017context} proposed Context Aware Correlation Filter tracking (CACF) framework where global context information is integrated within Scale Adaptive Multiple Feature (SAMF) \cite{li2014scale} as baseline tracker. The model is improved to compute high responses for targets, while close to zero responses for context information. 
The SAMF \cite{li2014scale} uses KCF as baseline and solves the scaling issue by constructing a pool containing the target at different scales. Bilinear interpolation is employed to resize the samples in the pool to a fixed size template.\\ 
Hu et al. \cite{hu2017manifold} proposed Manifold Regularized Correlation object-Tracking with Augmented Samples (MRCT-AS) to exploit the geometric structure of the target, and introduced a block optimization mechanism to learn manifold regularization. 
Unlike the KCF tracker, the MRCT-AS mines negative samples while maintaining a certain distance from the target. 
Labeled and unlabeled samples are augmented to construct Gram matrix with block circulant structure. 
A Gaussian kernal is used to construct kernel matrix. Laplacian regularized least squares \cite{belkin2006manifold} is employed to impose manifold structure on the learning model.
An affinity matrix is constructed from the similarity of samples using radial basic function to construct block circulant structural Laplacian matrix. The model has been optimized using a dialognalization method.
The objective of manifold regularization is to label unlabeled neighboring samples with the same labels.  The confidence map for unlabeled sample is computed from the learned model, and maximum response estimates the target position.\\
The Structuralist Cognitive model for Tacking (SCT) \cite{choi2016visual} divides the target into several cognitive units.
During tracking, the search region is decomposed into fixed-size grid map, and an individual Attentional Weight Map (AWM) is computed for each grid cell. The AWM is computed from the weighted sum of Attentional Weight Estimators (AWE). The AWE assigns higher weights to target grid and lower weights to background grid using a Partially Growing Decision Tree (PGDT) \cite{choi2015user}. Each unit works as individual KCF \cite{henriques2015high} with Attentinal CF (AtCF), having different kernel types with distinct features  and corresponding AWM.  
The priority and reliability of each unit are computed based on relative performance among AtCFs and its own performance, respectively. Integration of response maps of individual units gives target position.\\ 
Choi et al. proposed a Attentional CF Network (ACFN) \cite{choi2017attentional} exploits  target dynamic based on an attentional mechanism. An ACFN is composed of a CF Network (CFN) and Attentional Network (AN). The CFN has several tracking modules that compute tracking validation scores as precision. The KCF is used for each tracking module with AtCF and AWM. The AN selects tracking modules to learn target dynamics and properties. The AN consists of two sub networks i.e. Prediction Sub Network (PSN) and Selection Sub Network (SSN). Validation scores for all modules are predicted in PSN. The SSN chooses active tracking modules based on current predicted scores.
The target is estimated as that having the best response among the selected subset of tracking modules.\\ 
\subsubsection{Siamese Based Correlation Filter Trackers}
Recently, visual tracking via Siamese network has been used to handle  tracking challenges, including \cite{guo2017learning, bertinetto2016fully, valmadre2017end, tao2016siamese}. A Siamese network joins two inputs and produces a single output. The objective is to determine whether identical objects exist, or not, in the two image patches that are input to the network. The network measures similarity between the two inputs, and has the capability to learn similarity and features jointly. Bromley et al. \cite{bromley1994signature} and Baldi et al. \cite{baldi2008neural} first introduced the concept of Siamese network in their work on signature verification and fingerprint recognition, respectively.  Later, Siamese networks were used in many computer vision application, such as face recognition and verification \cite{schroff2015facenet}, stereo matching \cite{zbontar2015computing}, optical flow \cite{dosovitskiy2015flownet}, large scale video classification \cite{karpathy2014large} and patch matching \cite{zagoruyko2015learning}.\\ 
Fully convolutional Siamese networks (SiameseFC) \cite{bertinetto2016fully} solves the tracking problem using similarity learning that compares exemplar (target) image with a same-size candidate image, and yields high scores if the objects are the same. The SiameseFC algorithm is fully convolutional, and its ouptput is a scalar-valued score map that takes as input an example target and search patch larger than target predicted in the previous frame. 
The SiameseFC network utilizes a convolutional embedding function and a correlation layer to combine feature maps of the target and search patch. Target position is predicted by the position of maximum value in the score map. This  gives frame to frame target displacement.\\ 
Valmadre et al. \cite{valmadre2017end} introduced Correlation Filter Network (CFNet) for end-to-end learning of underlying feature representations through gradient back propagation. 
SiameseFC is used as base tracker, and CFNet is employed in forward mode for online tracking. During the online tracking of CFNet, target features are compared with the larger search area on new frame based on previously estimated target location. A similarity map is produced by computing the cross-correlation between the target template and the search patch.\\ 
The Discriminative Correlation Filters Network (DCFNet) \cite{wang17dcfnet}  utilizes lightweight CNN network with correlation filters to perform tracking using offline training. The DCFNet performs back propagation to learn the correlation filter layer using a probability heat-map of target position.\\ 
Recently, Guo et al. \cite{guo2017learning} presented Dynamic Siamese (DSaim) network that has the potential to reliably learn online temporal appearance variations. The DSaim exploits CNN features for target appearance and search patch. Contrary to the SiameseFC, the DSaim learns target appearance and background suppression from previous frame by introducing Regularized Linear Regression (RLR) \cite{scholkopf2002learning}. Target appearance variations are learned from first frame to current frame, while background suppression is performed by multiplying the search patch with the learned Gaussian weight map. The DSaim performs element-wise deep feature fusion through circular convolution layers to multiply inputs with weight map. Huang presented  EArly Stopping Tracker (EAST) \cite{huang2017learning} to learn polices using deep reinforcement learning and improving speedup while maintaining accuracy. The tracking problem is solved using Markov Decision Process (MDP). A RL agent makes decision based on  multiple  scales with an early stopping criterion. The objective is to find a tight bounding box around the target.\\
\subsubsection{Part Based Correlation Filter Trackers}
These kind of trackers learn target appearance in parts, unlike other CFTs where target template is learned as a whole. Variations may appear in a sequence, not just because of illumination and viewpoint, but also due to intra-class variability, background clutter, occlusion, and deformation. For example, an object may appear in front of the object being tracked, or a target may undergo non-rigid appearance variations.  There are many computer vision applications that use part-based techniques, such as object detection \cite{felzenszwalb2010object, forsyth2014object}, pedestrian detection \cite{prioletti2013part} and face recognition \cite{joseph2003holistic}.
Tracking algorithms \cite{vcehovin2011adaptive, ross2008incremental, yang2012online, liu2015real, sun2017non} have been developed to solve the challenges where targets are occluded or deformed in the sequences.\\
Liu et al. \cite{liu2015real} proposed Real time Part based tracking with Adaptive CFs (RPAC), which adds a spatial constraint to each part of object. 
During  tracking, adaptive weights as confidence scores for each part are calculated by computing sharpness of response map and Smooth Constraint of Confidence Map (SCCM). Response sharpness is calculated using Peak-to-Sidelobe Ratio (PSR),  while SCCM is defined by the temporal shift of part between two consecutive frames. Adaptive part trackers are updated for those parts whose weights are higher then a threshold value. A Bayesian inference theorem is employed to compute the target position by calculating the Maximum A Posteriori (MAP) for all the parts of object.\\
Liu at al. \cite{liu2016part} upgraded RPAC to RPAC+ based on Bayesian inference framework to track multiple object parts with CFs and adapt appearance changes from structural constrained mask using adaptive update strategy. The SCCM is used to select discriminative parts efficiently and suppress noisy parts. RPAC+ is improved by assigning proper weights to parts. Instead of tracking fix five parts, tacker accommodates various parts. RPAC+ begins with a large number of part models, then reduces to small number of trackers. During  tracking, parts are sorted in descending order based on their confidence scores. Overlapping scores are calculated for parts, and if two parts have greater then 0.5 score, then part with lower confidence score is discarded.\\
The Reliable Patch Tracker (RPT) \cite{li2015reliable} is based on particle filter framework which apply KCF as base tracker for each particle, and exploits local context by tracking the target with reliable patches. During tracking, the weight for each reliable patch is calculated based on whether it is a trackable patch, and whether it is a patch with target properties. The PSR score is used to identify patches, while motion information is exploited for probability that a patch is on target. Foreground and background particles are tracked along with relative trajectories of particles. A patch is discarded if it is no longer reliable, and re-sampled to estimate a new reliable patch. A new target position is estimated by aHough Voting-like strategy by obtaining all the weighted, trackable, reliable positive patches.
Recurrently Target attending Tracking (RTT) \cite{cui2016recurrently} learns the model by discovering and exploiting the reliable spatial-temporal parts using DCF. Quard-directional Recurrent Neural Network (RNNs) are employed to identify reliable parts from different angles as long-range contextual cues. Confidence maps from RNNs are used to weight adaptive CFs during tracking to suppress the negative effects of background clutter. 
Patch based KCF (PKCF) \cite{chen2016robust} is a particle filter framework to train target patches using KCF as base tracker. 
Adaptive weights as confidence measure for all parts based on the PSR score are computed. For every incoming frame,  new particles are generated from the old particle set, and responses for each template patch are computed. The PSR for each patch is calculated, and the particles with maximum weights are selected.\\ 
The Enhanced Structural Correlation (ESC) tracker \cite{chen2017visual} exploits holistic and object parts information. The target is estimated based on weighted responses from non-occluded parts. Colour histogram model, based on Bayes' classifier is used to suppress background by giving higher probability to objects. The background context is enhanced from four different directions, and is considered for the histogram model of the object's surroundings. The enhanced image is decomposed into one holistic and four local patches. The CF is applied to all image patches and final responses are obtained from the weighted response of the filters. Weight as a confidence score for each part is measured from the object likelihood map and the maximum responses of the patch. Adaptive CFs are updated for those patches whose confidence score exceeds a threshold value. Histogram model for object are updated if the confidence score of object is greater then a threshold value,  while background histogram model is updated on each frame.
Zuo et al.\cite{liu2016structural} proposed Structural CF (SCF) to exploit the spatial structure among the parts of an object in its dual form. The position for each part is estimated at the maximum response from the filter response map.  Finally, the target is estimated based on the weighted average of translations of all parts, where the  weight of individual part is the maximum score on the response map.\\
\subsection{Patch Learning Based Tracker}
Patch learning-based trackers exploit both target and background patches. A tracker is trained on positive and negative samples. The Model is tested on number of samples, and the maximum response gives the target position.\\
Zhang et al. \cite{zhang2016robust} proposed Convolutional Networks without Training (CNT) tracker that exploits the inner geometry and local structural information of the target. The CNT algorithm is an adaptive algorithm based on particle filter framework in which appearance variation of target is adapted during the tracking. CNT employs a hierarchical architecture with two feed forward layers of convolutional network to generate an effective target representation. In the CNT, pre-processing is performed on each input image where image is warped and normalized. The normalized image is then densely sampled as a set of overlapping local image patches of fixed size, also known as filters, in the first frame. After pre-processing, a feature map is generated from a bank of filters selected with k-mean algorithm. Each filter is convolved with normalized image patch, which is known as simple cell feature map. In second layer, called complex cell feature map, a global representation of target is formed by stacking simple cell feature map which encodes local as well as geometric layout information. 
Exemplar based Linear Discriminant Analysis (ELDA) \cite{gao2017robust} employs LDA to distinguish the target from the background. ELDA takes one positive sample at current target position and negative samples from the background.  ELDA has object and background component models. The object model consists of two models: a long-term  and a short-term  model. The long-term model corresponds to the target template from the first frame, while target appearance in a sort time window corresponds to the short-term model.  The background models also consists of two models: one offline and an online background models. The offline background model is trained on large number of negative samples from natural images, while the online is built from negative samples around the target. 
The ELDA tracker is comprised of a long-term detector and a  short-term detector. Target location is estimated from the sum of long-term and  weighted sum of   short-term detection scores. 
ELDA has been enhanced by integration with CNN, and named as Enhanced CNN Tracker (ECT) \cite{gao2016enhancement}.\\ 
A Multi-Domain Network (MDNet) \cite{nam2016mdnet}  consists of shared layers (three convolutional layers and two fully-connected layers) and one domain-specific fully connected layer. Shared layers exploit generic target representation from all the sequences, while domain specific layer are responsible for identification of target using binary classification for a specific sequence.
During  online tracking, the domain specific layer is initialized at the first frame. Samples are generated based on previous target location, and a maximum positive score yields the new target position. Weights of the three convolutional layers are fixed while  weights of three fully connected layers are updated for short- and long-term update. Long-term update is performed after a fixed long-term interval from positive samples. The short-term update is performed whenever tracking fails and the weights of fully-connected layers are updated using positive and negative samples from the current short term interval. 
A bounding box regression  model \cite{girshick2014rich} is also used to adjust the predicted target position in the subsequent frames.\\
A Structure Aware Network (SANet) \cite{fan2016sanet} exploits the target's structural information based on particle filter framework. 
The structure of target is encoded by a RNN via an undirected cyclic graph. 
SANet's architecture is similar to MDNet architecture, with the difference of each pooling layer being followed by a recurrent layer.  A Skip concatenation method is adopted to fuse output features from pooling and recurrent layers.\\ 
Han et al. \cite{han2017branchout} presented BranchOut algorithm, which uses MDNet as base tracker.  The BranchOut architecture comprises of three CNN layers and multiple fully-connected layers as branches. Some branches consists of one fully-connected layer, while some others have two fully-connected layers. During  tracking, a random subset of branches is selected by Bernoulli distribution to learn target appearance.\\
The Biologically Inspired Tracker (BIT) \cite{cai2016bit} performs tracking like ventral stream processing. The  BIT  tracking framework consists of an appearance model and a tracking model.  The appearance model consists of two units, classical simple cells (S1) and cortical complex cells (C1). A S1 is responsible to exploiting colour and texture information, 
while a C1 performs pooling and combining of color and texture features to form complex cell. 
The tracking model also have two units, a view-tuned learning (S2) unit and a task dependent learning (C2) unit. 
S2 computes response map  by performing convolution between the input features, and the target  and response maps are fused via average pooling. The C2 unit then computes new target position by applying CNN classifier.\\
An Action-Decision Network (ADNet) \cite{yun2017adnet} controls sequential actions (translation, scale changes, and stopping action) for tracking using deep Reinforcement Learning (RL). 
The network consists of three convolutional layers and three fully-connected layers.
An ADNet is defined as an agent with the objective to find target bounding box. The agent is pretrained to make decision about target's movement from a defined set of actions. During  tracking, target is tracked based on estimated action from network at the current tracker location. Actions are repeatedly estimated by agent unless reliable target position is estimated. Under the RL, the agent gets rewarded when it succeeds in tracking the target, otherwise, it gets penalized.\\ 
Zhang et al.\cite{zhang2017deep} presented Deep RL Tracker (DRLT), which consists of observations and recurrent networks. An observation network is responsible for computing deep features, while a recurrent network computes hidden states from deep features and previous hidden states. The target position is estimated from a  newly-generated hidden state. 
During offline training, the agent receives a reward for each time step, with the objective is to maximize the reward.  The tracker chooses several consecutive frames and computes features, hidden states and outputs. A set of target positions are estimated for selected frames, and a reward for each estimation is calculated. Network parameters are updated based on sum of rewards.\\ 
The Oblique Random forest (Obli-Raf) \cite{Zhang2017CVPR}  exploits geometric structure of the target. During  tracking, sample patches are drawn as particles, based on estimated target position on previous frame. Extracted particles are fed  to an oblique random forest classifier. Obli-Raf uses proximal support vector machine (PSVM) \cite{mangasarian2001proximal} to obtain the hyperplane from data particles in a semi-supervised manner to recursively cluster sample particles. Particles are classified as target or background, based on votes at each leaf node of the tree. The particle with the maximum score will be considered as newly-predicted target position. If the number of votes are less then a predefined threshold, then a new set of particle samples are drawn from the estimated target position. The model is updated if the maximum number of votes are greater then a threshold value, otherwise the previous model is retained.\\ 
A Temporal Spatial Network (TSN) \cite{teng2017robust} exploits the spatial and temporal features to refine  predicted target location. TSN is composed of three nets: (1) Feature Net (FN) to generate deep features, (2) Temporal Net (TN) to compute the similarity between current frame and historic feature maps and (3) a Spatial Net (SN)  to refine the target location. Training samples are cropped at the first frame of the sequence, and TN and SN are trained. During  tracking, samples at previous estimated target location are cropped and forwarded to FN to compute features. The TN estimates the similarities between candidate feature map and template feature map. Finally, SN gives the target position corresponding to the maximum response location.\\
Zhu at al.\cite{zhu2016beyond} proposed Edge Box Tracker (EBT) to perform global search to locate a target without considering a specific search window. The EdgeBox \cite{zitnick2014edge} is used for object proposal, as the object bounding box is based on likelihood of object (objectness) and Structured Support Vector Machine (SSVM) is employed for classification.\\
Deep Relative Tracking (DRT) \cite{gao2017deep} is based on particle filter framework that introduces a relative loss layer to model relative information among patches. A DRT network consists of five convolutional, five fully-connected layers, and one relative loss layer. Training of network involves two side of networks, with shared weights that take as input the two patches and the overlap score. Input images are divided into six subsets, depending upon their overlap ratio. Image pairs are ordered in different subsets such that similar image pairs are placed at the last.
During  tracking, one side of network is used to predict relative score to estimate the target position.\\
Li at al.\cite{li2016deeptrack} proposed a Deep Tracker (DeepTrack) to learn structural and truncated loss function to exploit target appearance cues. Its architecture takes three image cues, and is composed of two convolutional, two fully-connected layers, and a fusion layer to fuse all features from different image cues. During  tracking, the target is estimated and training samples are generated around estimated target. The training sample pool for temporal target appearance adaptation increases gradually, depending upon quality of samples. The quality of training samples is computed using conditional probability. CNN weights are updated in minibatch from training sample pool if training loss is greater then threshold.\\
\subsection{Multiple Instance Learning Based Tracker}
Usually, visual trackers  update  appearance model after a regular interval of time. Training samples play a crucial role to update. One of the most common approach is to take one positive example at newly-estimated location, and negative examples around neighborhood of current position. If predicted location is not precise, the model may degrade over time and  cause drift problem. Another approach is to take multiple positive examples along with negative samples,  so the model does not lose its discriminative ability. Therefore, there is a need to crop samples in a more expressive way to tackle those problems.
Dietterich et al. \cite{dietterich1997solving} introduced Multiple Instance Learning (MIL).  In MIL, training examples are presented in bags instead of individual,  and the  bags, not the instances, are labeled. A bag is labeled positive if it has at-least one positive sample in it and negative bag contains all negative samples. Positive bag may contain positive and negative instances. During training in MIL, label for instances are unknown but bag labels are known. In the MIL tracking framework, instances are used to construct weak classifiers, and a few instances are selected and combined to form a strong classifier. 
There are many computer visions tasks where MIL is being used for example object detection \cite{zhang2006multiple}, face detection \cite{guillaumin2010multiple} and action recognition \cite{ali2010human}.
Various researcher have employed MIL to track targets \cite{babenko2011robust, wang2017patch, yang2017visual, sharma2017mil, xu2015robust, abdechiri2017visual}.\\
Babenko et al. \cite{babenko2011robust} proposed a novel MIL Boosting (MILBoost) algorithm to label ambiguity of instances using Haar features. A strong classifier is trained to detect a target by choosing  weak classifiers. A weak classifier is computed using log  odds ratio in a Gaussian distribution. A Noisy-OR model is used to compute the bag probabilities. MILBoost selects weak classifiers from the candidate pool based on maximum log likelihood of bags. Finally, new target position is estimated based on strong classifier as the weighted sum of weak classifiers.\\
Xu et al. \cite{xu2015robust} proposed an MIL framework that uses Fisher information using MILBoost (FMIL) to select weak classifiers.
Uncertainty is measured from unlabeled samples in fisher information criterion \cite{cover2006elements}. Feature subsets are selected to maximize the fisher information of the bag.
Abdechiri et al. \cite{abdechiri2017visual} proposed Chaotic theory in MIL (CMIL). Chaotic representation exploits complex local and global target information. HOG and Distribution Fields (DF) features with optimal dimension are used for target representation.
Chaotic approximation is used in the discriminative classifier.
The significance of instances are computed using fractal dimensions of state space and position distance simultaneously. The chaotic model is learned to adapt dynamic of target through chaotic map to maximize likelihood of bags.
To encode chaotic information, state space is reconstructed. An image patch is embedded into state space by converting it into a vector form and normalizing it with a mean equal to  0 and variance equal to 1. Taken's embedding theory generate a multi-dimensional space map from one-dimension space. The minimum time delay and the embedding dimension are predicted by false nearest neighbours to reduce dimensionality for state space reconstruction. Finally, GMM is imposed to model state space.\\
Wang et al. \cite{wang2017patch} presented Patch based MIL (P-MIL) that decomposes the target into several blocks. The MIL for each block is applied, and the P-MIL generates strong classifiers for target blocks. The average classification score, from classification scores for each block, is used to detect whole target. Sharma and Mahapatra \cite{sharma2017mil} proposed a MIL tracker based on maximizing the Classifier ScoRe (CSR) for feature selection. The tracking framework computes Haar-features for target with kernel trick, half target space, and scaling strategy.\\ 
Yang et al.\cite{yang2017visual} used Mahalanobis distance to compute the instance significance to bag probability in a MIL framework, and employed gradient boosting to train classifiers. During tracking, a coarse-to-fine search strategy is applied to compute instances. The Mahalanobis distance is used to define the importance between instances and bags. Discriminative weak classifiers are selected by maximizing the margin between negative and positive bags by exploiting the average gradient and average classifier strategy.\\
\subsection{Sparsity Based Tracker}
Sparse representation has been used by statistical signal processing, image processing, and computer vision communities for a number of applications including image classification \cite{romero2016unsupervised}, object detection \cite{peng2017salient}, and face recognition \cite{lou2016graph}.
The objective is to discover an optimal representation of the target which is sufficiently sparse and minimizes the reconstruction error. 
Mostly sparse coding is performed by first learning a dictionary. Assume $\textbf{X} =[x_1,...,x_N] \in \mathcal{R}^{m \times n}$ represents  gray scale images $x_i \in \mathcal{R}^m$.  A dictionary $\textbf{D}=[d_1,...,d_k] \in \mathcal{R}^{m \times k}$ is learned on $\textbf{X}$ such that each image in \textbf{X} can be sparsely represented by a linear combination of items in \textbf{D}: $x_i=\textbf{D}\alpha_i$, where $\alpha_i=[\alpha_1,...,\alpha_k] \in \mathcal{R}^{k}$ denotes the spares coefficients. 
When $k>r$, where $r$ is the rank of $\textbf{X}$, then dictionary $\textbf{D}$ is overcomplete. For a known $\textbf{D}$, a constrained minimization using $\ell_1-$norm is often applied to find $\alpha$ for sufficiently sparse solution: 
\begin{equation}\label{SC:3}
\alpha_i^* \equiv \underset{\alpha_i}{\text{arg min}}\frac{1}{2}\parallel x_i-\textbf{D}\alpha_i \parallel_2^2 + \lambda \parallel \alpha_i \parallel_1,
\end{equation}
where $\lambda$ gives relative weights to the sparsity and reconstruction error. Dictionary \textbf{D} is learned in such  a way that all images in $\textbf{X}$ can be sparsely represented with a small error. Dictionary $\textbf{D}$ is learned to solve following optimization problem:
\begin{equation}\label{SC:4}
\{\alpha^*, \textbf{D}^*\} \equiv \underset{\textbf{D},\alpha}{\text{minimize}}  \sum_{i=1}^N \parallel \textbf{X}-\textbf{D}\alpha \parallel_2^2 + \lambda \parallel \alpha \parallel_1,
\end{equation}
There are two alternative phases for dictionary learning. In the first phase, \textbf{D} is assumed to be fixed and the coefficients $\alpha$ are computed, while in the second phase, dictionary $\textbf{D}$ is updated and $\alpha$ is assumed to be fixed.  In visual object tracking, objective of dictionary learning is to distinguish a target from the background patches by sparsely encoding target and background coefficients.\\ 

Structural sparse tracking \cite{zhang2015structural} (SST) is based on particle filter framework which exploits intrinsic relationship of local target patches and global target to jointly learn sparse representation. The target location is estimated from target dictionary templates and corresponding patches having a maximum similarity score from all the particles. The model is constructed on a number of particles representing target, and each target representation is decomposed into patches, and dictionary is learned. The patch coefficient is learned such that it minimizes the patch reconstruction error.\\
Guo et al. \cite{guo2017visual} computed  weight maps to exploit  reliable target structure information. 
Traditional sparse representation is integrated along with reliable structural information.
A reliable structural constraint is imposed by the weight maps to preserve the target and background structure. Target template coefficients and weight maps template coefficients are optimized (minimized) together using Accelerated Proximal Gradient (APG) method. The pyramidal Lucas-Kanade \cite{bouguet2001pyramidal} is used to construct weight map. Using  a Bayesian filtering framework, target is estimated using maximum likelihood from the estimated object state for all the particles.\\
Yi et al. \cite{yi2017visual} proposed Hierarchical Sparse Tracker (HST) to integrate the discriminative and generative models. The proposed appearance model is comprised of Local Histogram Model (LHM), Weighted Aligment Pooling (WAP), and Sparsity based Discriminant Model (SDM). 
LHM encodes the spatial information among  target parts while the WAP assigns weights to local patches based on  similarities between target and candidates. The target template sparse representation is computed in SDM. Finally, candidate with the maximum score from LHM, WAP, and SDM determines the new target position.\\
Context aware Eclusive Sparse Tracker (CEST) \cite{zhang2016robust} exploits context information based on particle filter framework. The CEST represents particles as a linear combination of dictionary templates. Dictionary is modeled as groups containing templates as  target, occlusion or noise, and context. Inter- and intra-type sparsity is hold for each group. An efficient Accelerated Proximal Gradient (APG) method is used to learn particle representations.\\
\subsection{Superpixel Based Tracker}
In image processing, the pixel is the smallest physical controlable element. 
Pixels represent the colour intensities of the objects in images. As the object appearance changes, pixel information also changes, thus pixels are not the best way to represent object. However, superpixels give perceptual information about rigid structure of pixel grid. Superpixels represent the group of pixels having identical pixel values \cite{achanta2012slic}. 
A superpixel based representation got much attention by computer vision community for object recognition \cite{fulkerson2009class}, human detection \cite{mori2004recovering}, activity recognition \cite{wu2007scalable}, and image segmentation \cite{ achanta2012slic}. Numerous tracking algorithms have been developed using superpixels \cite{huang2017structural, wang2017constrained, wang2017two, jingjing2016tracking, li2014object}.\\
The tracker introduced by Jingjing et al.\cite{jingjing2016tracking}  is based on Bayesian framework. The model is trained over target and background superpixels. Superpixels are divided into clusters using mean shift algorithm. Weights for each cluster is computed and sorted. 
The superpixels score map is calculated from three factors: the distance between the superpixel and the cluster center it belongs to, cluster weight and  label, and whether the cluster belongs to target or background region. 
For every new frame, superpixels are computed around the surrounding region of target based on previous frame. Highest superpixel score estimates the target center on current frame.\\
The Constrained Superpixel Tracking (CST) \cite{wang2017constrained} algorithm employs graph labeling to integrate spatial smoothness, temporal smoothness, and appearance fitness constraints. 
Spatial smoothness is enforced by exploiting the latent manifold structural using unlabeled and labeled superpixels. Optical flow is used for the temporal smoothness to impose short-term target appearance, while appearance fitness servers as long-term appearance model to enforce objectness.
Structural manifold ranking \cite{zhou2004ranking} is used to label superpixels where the affinity matrix contains the penalty weights of two similar superpixels. 
For temporal smoothness, similarity between two superpixels is computed via optical flow by Lucas and Kanade \cite{lucas1981iterative} and affinity matrix is defined by similarity between two consecutive frame superpixels. Finally, a random forest tree is trained to classify target superpixels. 
During tracking, HSI colour histogram features are used for spatial and temporal constraint, while RGB features are used for appearance fitness constraint. A new target center is estimated on the current frame with maximum posteriori estimation over all candidates superpixels.\\
Wang et al. \cite{wang2017two} presented a Bayesian tracking method at two-level superpixel appearance model. Object outliers are computed using Bilateral filter. 
The coarse-level appearance model computes few superpixels such a way that there will one superpixel in bounding box of target, and a confidence measure defines whether the superpixel belongs to target or background.  The  fine-level appearance model calculates more superpixels then coarse-level over the target region based on target location on previous frame to compute the confidence map. 
The confidence map is computed from colour similarity and the relative positions of superpixels to impose the structural information of superpixels.\\
The Structural Superpixel Descriptor (SSD) \cite{huang2017structural} exploits the structural information via superpixels and preserves the intrinsic properties of target. It decomposes the target into hierarchy of different size superpixels and assign greater weights to superpixels closer to the object center. A particle filter framework is employed and background information is alleviated through adaptive patch weighting. AnSVM is used to estimate likelihood for candidate patches. 
Li et al. \cite{li2014object} used BacKGround (BKG) cues for tracking. During  tracking, the background is segmented excluding object for superpixel segmentation from previous frames. A weighted confidence map is computed based on difference between target and background using a PCA background colour model  from the k previous frames. Target position is estimated based on the candidate with the maximum weighted confidence score.\\ 
\subsection{Graph Based Tracker}
Graph represent suitable models to solve many computer vision problems \cite{deo2017graph}. Graph theory has many applications such as object detection \cite{deo2017graph, filali2016multi}, human activity recognition \cite{singh2017graph, li2016multiview}, and face recognition \cite{meena2017improved}. Generally, graph-based trackers use superpixels and node to represent the object  appearance, while edges  represent the inner geometric structure. Another strategy being used in graph-based trackers is to construct graphs among the parts of objects in different frames.\\
Graph Tracker (Gracker) \cite{wang2017gracker} uses undirected graphs to model planar objects and exploits the relationship between local parts. Search region is divided into grids, and a graph is constructed where vertices represents key points of maximum response using SIFT for each grid, and edges are constructed from Delaunary triangulation \cite{lee1980two}. During  tracking, geometric graph-matching is performed to explore optimal correspondence between model graph and candidate graph by computing affinity matrix graphs. Target is estimated at MAP estimation. 
Reweighted Random Walks for graph Matching (RRWM) \cite{cho2010reweighted} is used to refine matched graph.\\
Du et al. \cite{du2016online} proposed a Structure Aware Tracker (SAT) that constructs hypergraphs to exploits higher order dependencies in temporal domain. A SAT uses frame buffer to collect  candidate parts  from each frame in frame buffer by computing superpixels. A graph cut algorithm is employed to minimize the energy to produce the candidate parts. A structure-aware hyper graph is constructed with nodes representing candidate parts, while hyper edges denotes relationship among parts. 
A subgraph is built by grouping superpixels considering appearance and motion consistency of object parts across multiple frames. Finally, the target location and boundary is estimated by combining all the target parts using coarse-to-fine strategy.\\
A Geometric hyperGraph Tracker GGT \cite{du2016geometric} construct geometric hpyergraphs by exploring geometric relationships and learning to match the candidate part set and target part set. A geometric hypergraph is constructed from the superixels where vertices are correspondence hypothesis (possible correspondence between two parts sets with an appearance constraint) while edges constitute the geometric relationship within the hypothesis. During  tracking, reliable parts are extracted with high confidence to predict target location.  Reliable parts are the correspondence hypotheses learned from the matched target and candidate part sets.\\
The Tree structure CNN (TCNN) \cite{nam2016modeling} tracker employed CNN to model target appearance in tree structure. 
Multiple hierarchical CNN-based target appearance models are used to build a tree where vertices are CNNs and edges are relations among CNNs. Each path of tree maintains a separate history for target appearance in an interval. During tracking, candidate samples around the target location estimated in the previous frame are cropped.
Weighted average scores from multiple CNNs are used to compute abjectness for each sample. Reliable patch along the CNN defines the weight of CNN in the tree structure. The maximum score from multiple CNNs is used to estimate target location.
A bounding box regression \cite{girshick2014rich} method is also applied to enhance the estimated target position in the subsequent frames.\\
An Absorbing Markov Chain Tracker (AMCT) \cite{yeo2017superpixel} recursively propagates  the predicted segmented target in subsequent frames. 
AMC has two states: an absorbing and a transient state. In an AMC, any state can be entered to absorbing state, and once entered, cannot leave, while other states are transient states.
An AMC graph is constructed between two consecutive frames based on superpixels, where vertices are background superpixels (represents absorbing states) and target superpixels (transient states). Edges weights are learned from support vector regression to distinguish foreground and background superpixels. Motion information is imposed by spatial proximity using inter-frame edges.
The target is estimated from the superpixel components belonging to the target after vertices have been evaluated against the absorption time threshold.\\
\subsection{Siamese Network Based Tracker}
Siamese network perform tracking based on matching mechanism. The learning process exploits the general target appearance variations. Siamese network-based trackers match target templates with candidate samples to yield the similarities between patches. Basics of the Siamese is found with the discussion of Siamese-based CFT.\\
Siamese INstance Search (SINT) \cite{tao2016siamese} performs tracking using learned matching function, and finds best-matched patch between target template and candidate patches in new frames  without updating matching function. The SINT architecture have two steams: a query stream and search stream. Each steam is composed of five convolutional layers, three region-of-interest pooling layers, one fully-connected layer, and one fusion layer to fuse features.
Chen and Tao \cite{chen2017once} proposed  two flow CNN tracker called as YCNN that is learned end-to-end  shallow and deep features to measure the similarity  between the target patch and the search region. YCNN architecture has two flows: an object and search flow. Deep features obtained from object and search flows having three convolutional layers are concatenated, and are fed to two fully-connected layers and then to output layer.\\ 
Held et al.\cite{held2016learning} proposed Generic Object Tracking Using Regression Network (GOTURN) to exploit generic object appearance and motion relationships. Target and search regions are fed to five individual convolutional layers. Deep features from two separate flows are fused into shared three sequential fully-connected layers. 
GOTURN is a feed-forward offline tracker that does not require fine-tuning, and directly regresses target location.\\
Reinforced Decision Making (RDM) \cite{choi2017visual}  makes decision to select a template. A RDM model is composed of two networks: matching and policy networks. Prediction heatmaps are generated from the matching network,  while the policy network is responsible for producing normalized scores from prediction heatmaps. During tracking, a search patch is cropped from the target estimated in the previous frame and fed to matching networks along with target templates to produce prediction maps. Normalized scores are then produced by the policy network from prediction maps. The target is estimated at the maximum score of prediction map. The matching network consists of three shared convolutional layers and three fully-connected layers.  Features from shared convolutional layers are fused into fully-connected layers to produce prediction map. The policy network contains two convolutional and two fully-connected layers that make decisions about a reliable state using RL.\\
\subsection{Part Based Tracker}
Part based modeling have been activity used in non-CFTs to handle deformable parts of objects. There are many state-of-the-art techniques to perfrom object detection \cite{ouyang2017deepid}, action recognition \cite{du2016representation}, and face recognition \cite{zhang2016face} using parts. In part-based modeling, local parts are utilized to model tracker \cite{zhang2017adaptive, yao2017part, wang2016part, li2017visual}.\\
The Part based Tracker (PT) proposed by Yao et al.\cite{yao2017part} used latent variables to model unknown object parts. An object is decomposed into parts,nand each part is associated with adaptive weight. Offsets in the vicinity of the part are latent variables. A structural spatial constraint is also applied to each part using minimum spanning tree where vertices are parts and edges defines the consistent connection. A weight is assigned to each edge corresponding to Euclidean distance between two parts. 
Online latent structured learning using online Pegasos \cite{shalev2011pegasos} is performed for global object and its parts. 
During tracking, the maximum classification scores of object and parts estimates the new target position.\\
Li et al.\cite{li2017visual} used local descriptors to explore parts, and position relationship among parts. The target is divided into non overlapping parts. A pyramid  having  multiple local covariance descriptors is fused using max pooling depicting target appearance. Parts are modeled using star graph and central part of target representing central node.  Parts for all positions are selected from candidate pool and template parts by solving linear programming problem. During tracking, target is estimated from selected patches using weighted voting mechanism based on relationship between centre patch and surrounding patch.\\
A Part-based Multi-Graph Ranking Tracker \cite{wang2016part} PMGRT constructs graphs to rank local parts of a target. Multiple graphs are build from different part samples with various features. A weight is allocated to each graph. 
An affinity matrix is constructed based on multiple parts and feature types. Augmented Lagrangian formulation is optimized to select parts associated with confidence. Target is estimated from the parts having highest ranking.\\
Adaptive Local Movement Modeling (ALMM) \cite{zhang2017adaptive} improved the trackers by exploiting the local movements of object parts. Image patches positions are estimated using base trackers (Struck \cite{hare2016struck}, CT \cite{zhang2014fast}, STC \cite{zhang2014fast}  ) that represent target patch appearance, and patches are improved using GMM to prune out drifted patches.
GMM is employed to model the parts movement based on displacement of parts center to the global object center. Each patch is allocated a weigh based on motion and appearance for better estimation. The target position is estimated from a strong tracker by combining all parts trackers in a boosting framework. 
\begin{figure*}[]
\centering
\includegraphics[scale=0.5]{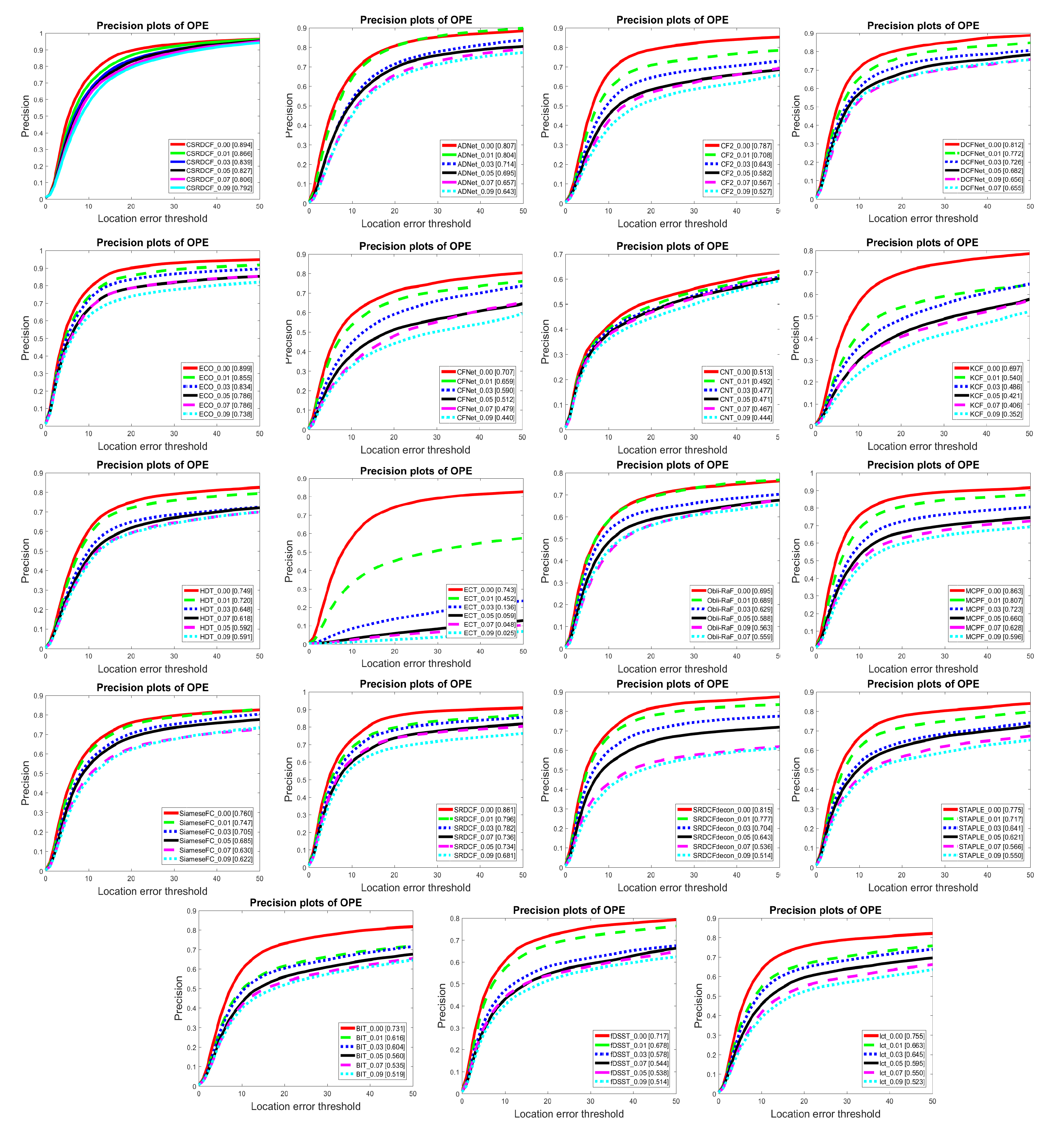}
\caption{Distance precision for CSRDCF \cite{Lukezic_CVPR_2017}, ADNet \cite{yun2017adnet}, CF2\cite{ma2015hierarchical}, DCFNet \cite{wang17dcfnet}, ECO\cite{DanelljanCVPR2017}, CFNet \cite{valmadre2017end}, CNT\cite{zhang2016robust}, KCF \cite{henriques2015high}, HDT\cite{qi2016hedged},  ECT \cite{gao2016enhancement}, Obli-Raf \cite{Zhang2017CVPR}, MCPF \cite{Zhang_2017_CVPR}, SiameseFC \cite{bertinetto2016fully}, SRDCF \cite{danelljan2015learning}, SRDCFdecon \cite{danelljan2016adaptive}, STAPLE\cite{bertinetto2016staple}, fDSST\cite{danelljan2017discriminative}, LCT \cite{ma2015long} and BIT \cite{cai2016bit} over OTB2015 benchmark \cite{wu2015object} using one-pass evaluation (OPE) with additive white Gaussian noise with zero mean and varying variance. The legend contains score at a threshold of 20 pixels for each tracker.}
\label{P_track}
\end{figure*}
\begin{figure}[t]
\centering
\includegraphics[width=\columnwidth]{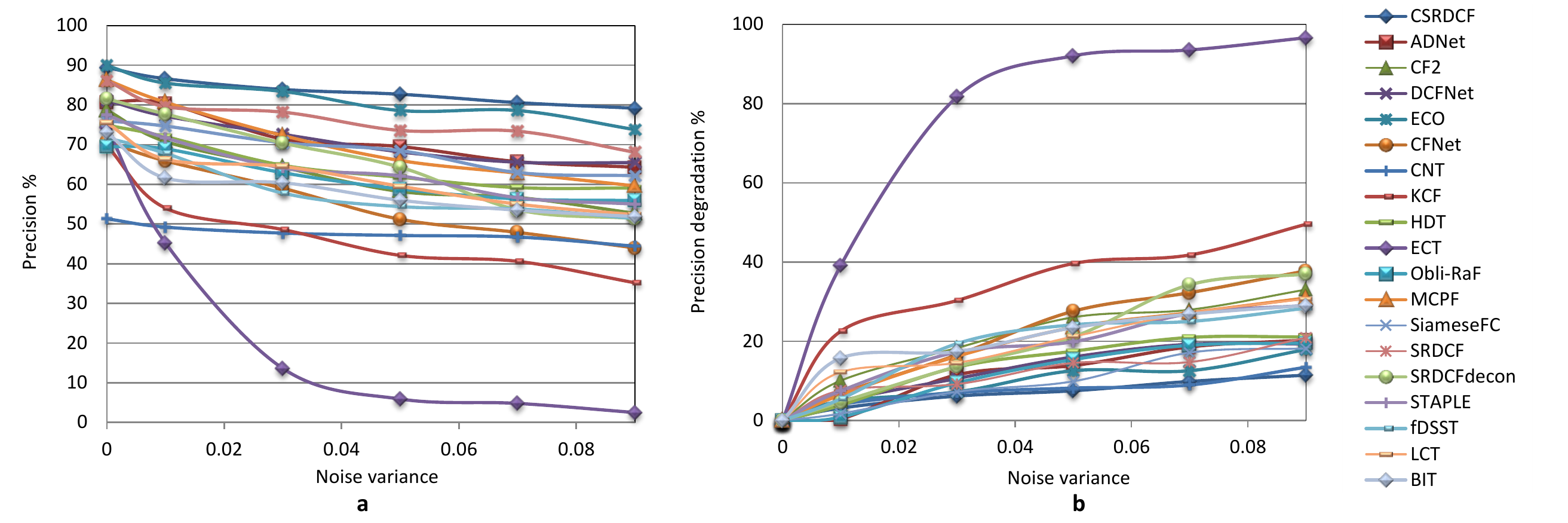}
\caption{Percentage of overall precision (a) and  percentage of precision degradation (b) plots for BIT CSRDCF \cite{Lukezic_CVPR_2017}, ADNet \cite{yun2017adnet}, CF2\cite{ma2015hierarchical}, DCFNet \cite{wang17dcfnet}, ECO\cite{DanelljanCVPR2017}, CFNet \cite{valmadre2017end}, CNT\cite{zhang2016robust}, KCF \cite{henriques2015high}, HDT\cite{qi2016hedged},  ECT \cite{gao2016enhancement}, Obli-Raf \cite{Zhang2017CVPR}, MCPF \cite{Zhang_2017_CVPR}, SiameseFC \cite{bertinetto2016fully}, SRDCF \cite{danelljan2015learning}, SRDCFdecon \cite{danelljan2016adaptive}, STAPLE\cite{bertinetto2016staple}, fDSST\cite{danelljan2017discriminative}, LCT \cite{ma2015long} and BIT \cite{cai2016bit} on series of Gaussian noise}
\label{P_Per_track}
\end{figure}
\begin{figure*}[]
\centering
\includegraphics[scale=0.5]{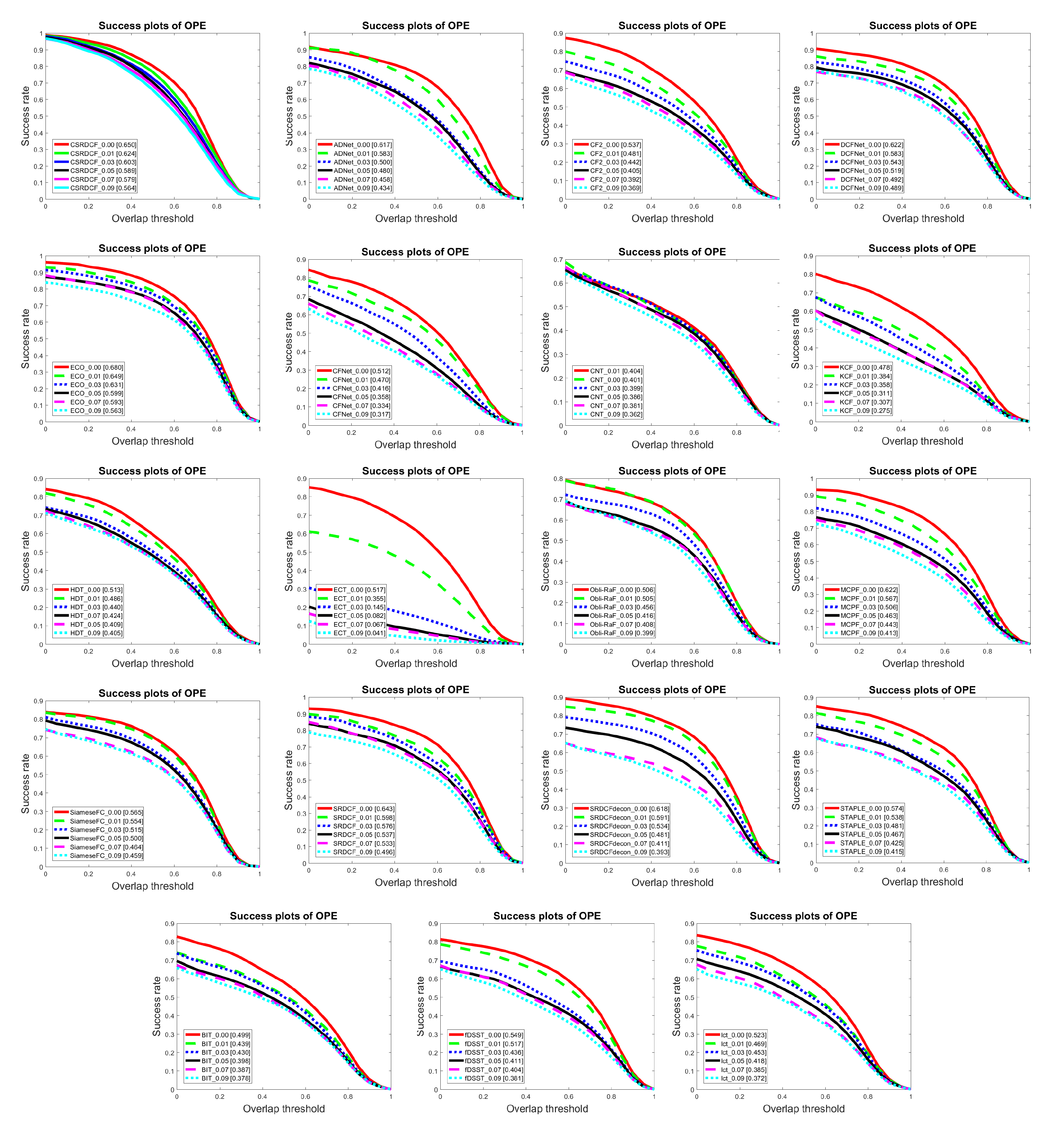}
\caption{Overlap success plots for CSRDCF \cite{Lukezic_CVPR_2017}, ADNet \cite{yun2017adnet}, CF2\cite{ma2015hierarchical}, DCFNet \cite{wang17dcfnet}, ECO\cite{DanelljanCVPR2017}, CFNet \cite{valmadre2017end}, CNT\cite{zhang2016robust}, KCF \cite{henriques2015high}, HDT\cite{qi2016hedged},  ECT \cite{gao2016enhancement}, Obli-Raf \cite{Zhang2017CVPR}, MCPF \cite{Zhang_2017_CVPR}, SiameseFC \cite{bertinetto2016fully}, SRDCF \cite{danelljan2015learning}, SRDCFdecon \cite{danelljan2016adaptive}, STAPLE\cite{bertinetto2016staple}, fDSST\cite{danelljan2017discriminative}, LCT \cite{ma2015long} and BIT \cite{cai2016bit} over OTB2015 benchmark \cite{wu2015object} using one-pass evaluation (OPE) with additive white Gaussian noise containing zero mean and varying variance. The legend contains overlap score at a threshold 0.5 for each tracker.}
\label{S_track}
\end{figure*}
\begin{figure}[t!]
\centering
\includegraphics[width=\columnwidth]{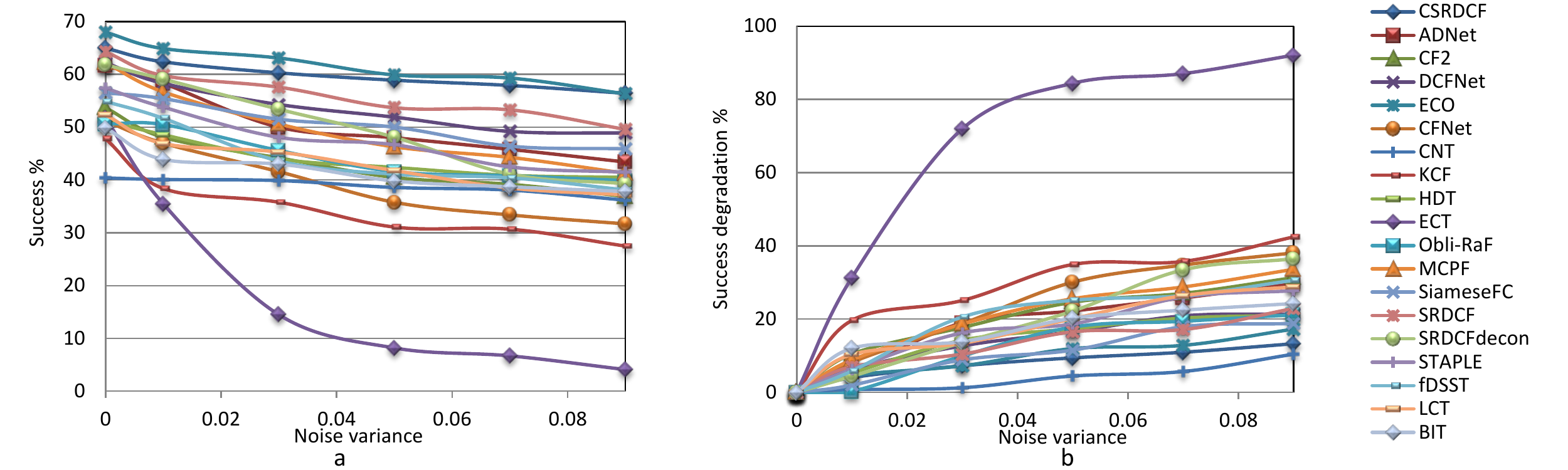}
\caption{Percentage of overall success (a) and  percentage of success degradation (b) plots for CSRDCF \cite{Lukezic_CVPR_2017}, ADNet \cite{yun2017adnet}, CF2\cite{ma2015hierarchical}, DCFNet \cite{wang17dcfnet}, ECO\cite{DanelljanCVPR2017}, CFNet \cite{valmadre2017end}, CNT\cite{zhang2016robust}, KCF \cite{henriques2015high}, HDT\cite{qi2016hedged},  ECT \cite{gao2016enhancement}, Obli-Raf \cite{Zhang2017CVPR}, MCPF \cite{Zhang_2017_CVPR}, SiameseFC \cite{bertinetto2016fully}, SRDCF \cite{danelljan2015learning}, SRDCFdecon \cite{danelljan2016adaptive}, STAPLE\cite{bertinetto2016staple}, fDSST\cite{danelljan2017discriminative}, LCT \cite{ma2015long} and BIT \cite{cai2016bit} on series of Gaussian noise}
\label{S_Per_track}
\end{figure}
\begin{figure*}[]
\centering
\includegraphics[scale=0.5]{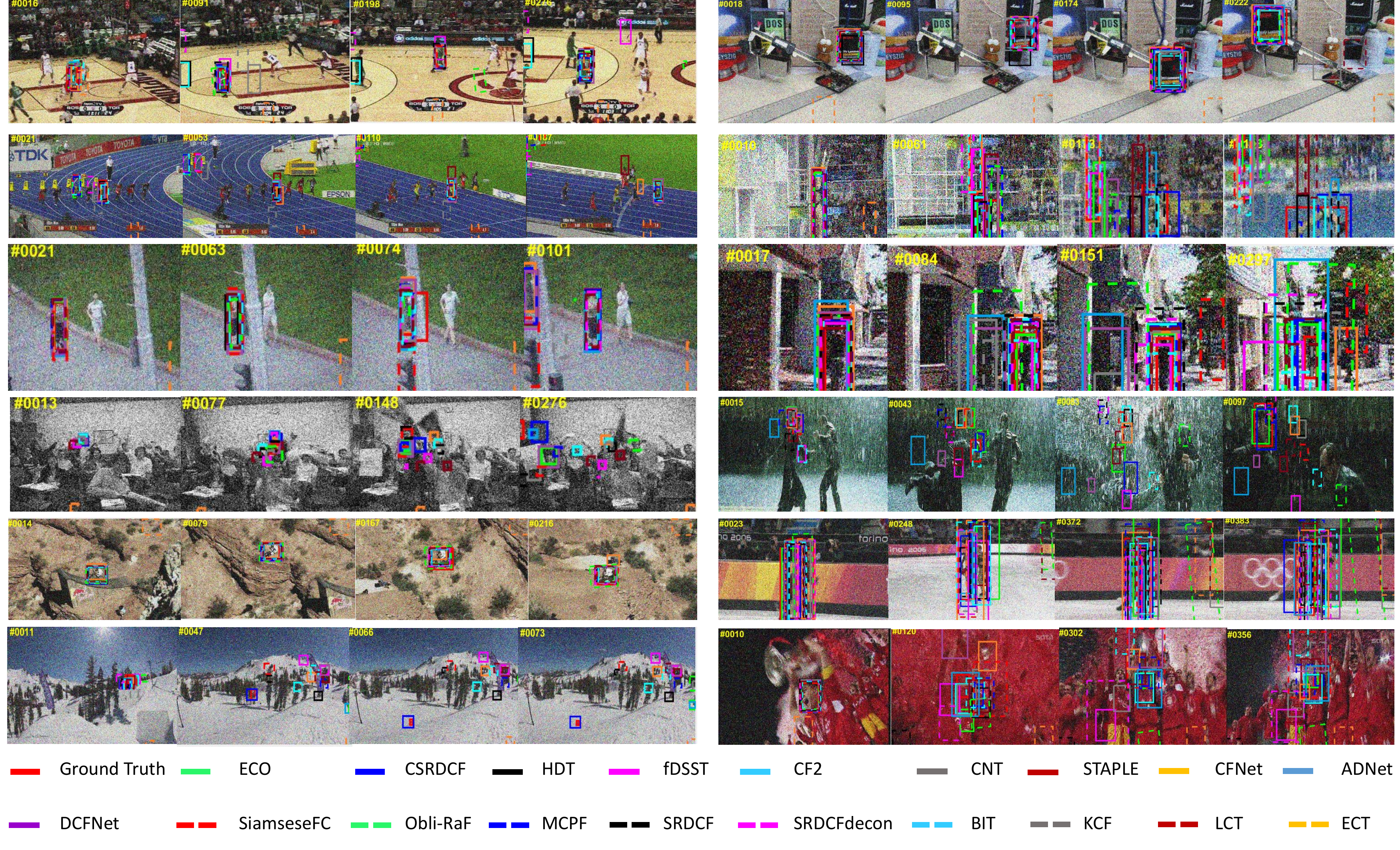}
\caption{Qualitative analysis of trackers ( CSRDCF \cite{Lukezic_CVPR_2017}, ADNet \cite{yun2017adnet}, CF2\cite{ma2015hierarchical}, DCFNet \cite{wang17dcfnet}, ECO\cite{DanelljanCVPR2017}, CFNet \cite{valmadre2017end}, CNT\cite{zhang2016robust}, KCF \cite{henriques2015high}, HDT\cite{qi2016hedged},  ECT \cite{gao2016enhancement}, Obli-Raf \cite{Zhang2017CVPR}, MCPF \cite{Zhang_2017_CVPR}, SiameseFC \cite{bertinetto2016fully}, SRDCF \cite{danelljan2015learning}, SRDCFdecon \cite{danelljan2016adaptive}, STAPLE\cite{bertinetto2016staple}, fDSST\cite{danelljan2017discriminative}, LCT \cite{ma2015long} and BIT \cite{cai2016bit}) on OTB2015\cite{wu2015object} containing additive Gaussian noise with zero mean and 0.05 variance on twelve challenging sequences (from left to right \textit{Basketball, Box, Bolt, Diving, Jogging-1, Human9, Freeman4, Matrix, MountainBike, Skating2-1, Skiing, and Soccer} respectively).}
\label{Q_I_05}
\end{figure*}
\begin{figure*}[]
\centering
\includegraphics[scale=0.8]{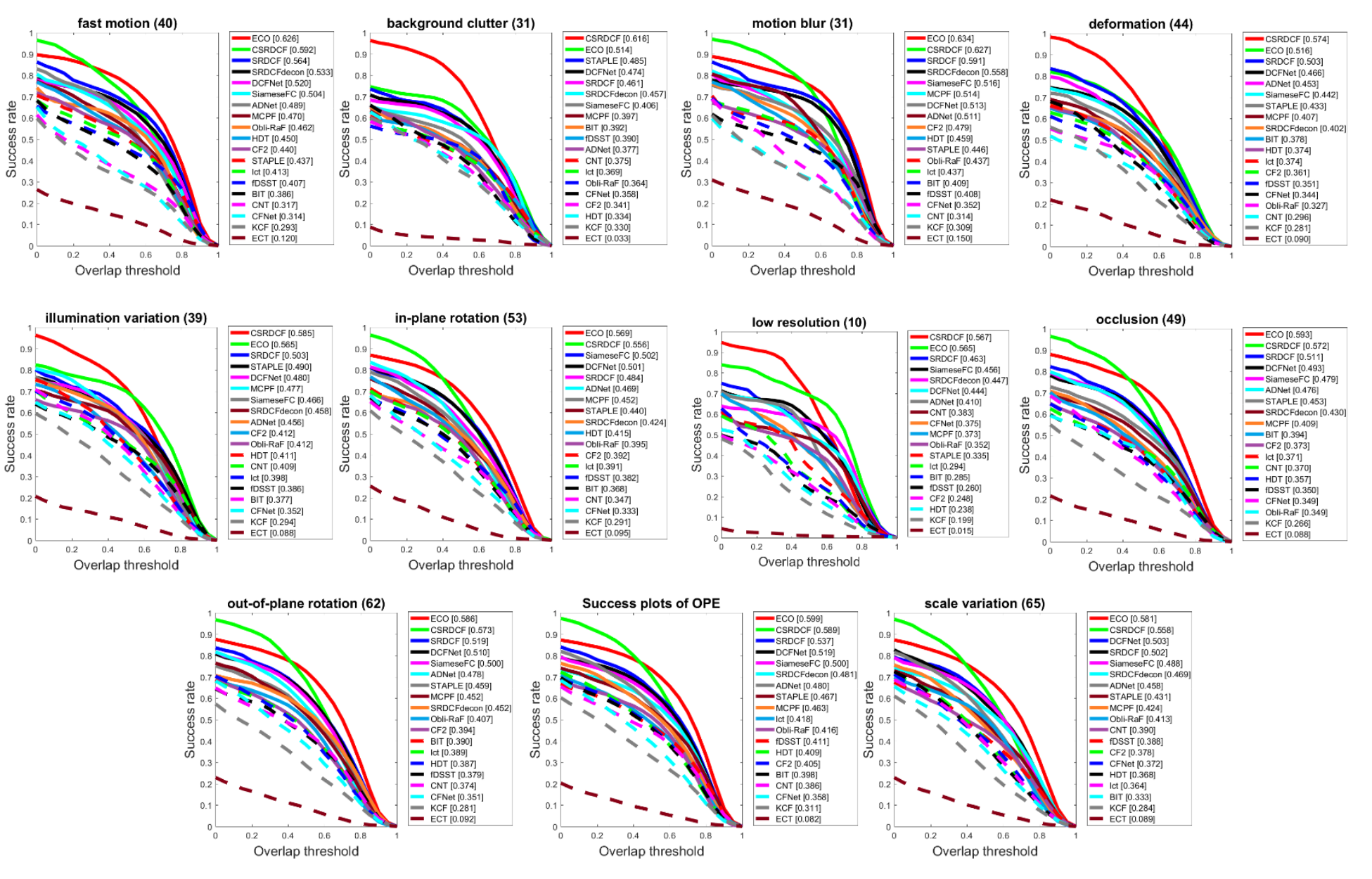}
\caption{Overlap success plots trackers ( CSRDCF \cite{Lukezic_CVPR_2017}, ADNet \cite{yun2017adnet}, CF2\cite{ma2015hierarchical}, DCFNet \cite{wang17dcfnet}, ECO\cite{DanelljanCVPR2017}, CFNet \cite{valmadre2017end}, CNT\cite{zhang2016robust}, KCF \cite{henriques2015high}, HDT\cite{qi2016hedged},  ECT \cite{gao2016enhancement}, Obli-Raf \cite{Zhang2017CVPR}, MCPF \cite{Zhang_2017_CVPR}, SiameseFC \cite{bertinetto2016fully}, SRDCF \cite{danelljan2015learning}, SRDCFdecon \cite{danelljan2016adaptive}, STAPLE\cite{bertinetto2016staple}, fDSST\cite{danelljan2017discriminative}, LCT \cite{ma2015long} and BIT \cite{cai2016bit}) on OTB2015\cite{wu2015object} containing additive Gaussian noise with zero mean and 0.05 variance over ten different challenges (background clutter, motion blur, deformation, illumination variation, in-plane rotation, low resolution, occlusion, fast motion, out of view, and scale variation). The legend contains overlap score at threshold 0.5 for each tracker.}
\label{S_C_05}
\end{figure*}
\begin{figure*}[]
\centering
\includegraphics[scale=0.8]{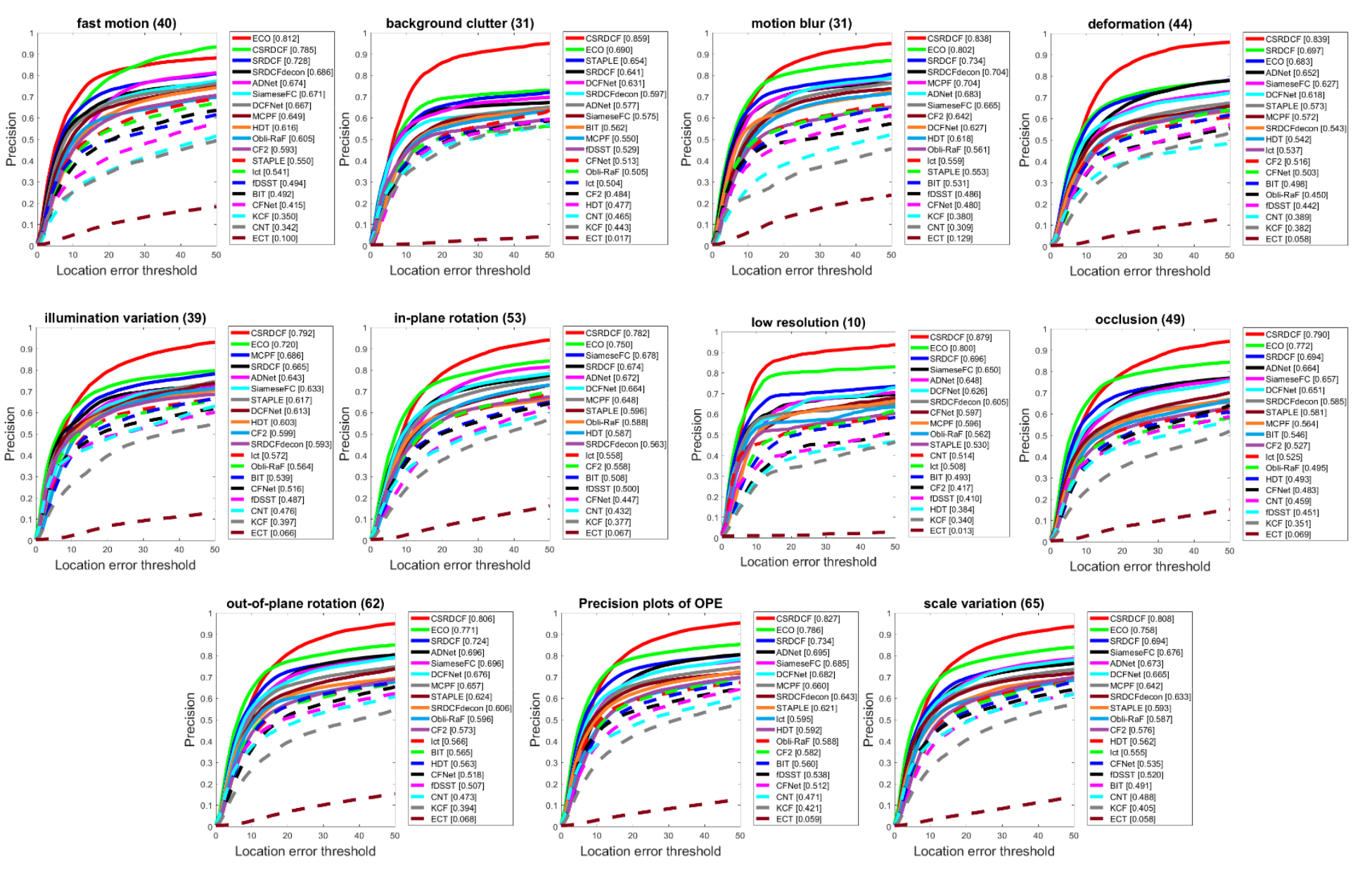}
\caption{Precision distance plot of trackers ( CSRDCF \cite{Lukezic_CVPR_2017}, ADNet \cite{yun2017adnet}, CF2\cite{ma2015hierarchical}, DCFNet \cite{wang17dcfnet}, ECO\cite{DanelljanCVPR2017}, CFNet \cite{valmadre2017end}, CNT\cite{zhang2016robust}, KCF \cite{henriques2015high}, HDT\cite{qi2016hedged},  ECT \cite{gao2016enhancement}, Obli-Raf \cite{Zhang2017CVPR}, MCPF \cite{Zhang_2017_CVPR}, SiameseFC \cite{bertinetto2016fully}, SRDCF \cite{danelljan2015learning}, SRDCFdecon \cite{danelljan2016adaptive}, STAPLE\cite{bertinetto2016staple}, fDSST\cite{danelljan2017discriminative}, LCT \cite{ma2015long} and BIT \cite{cai2016bit}) on OTB2015\cite{wu2015object} containing additive Gaussian noise with zero mean and 0.05 variance over ten different challenges (background clutter, motion blur, deformation, illumination variation, in-plane rotation, low resolution, occlusion, fast motion, out of view, and scale variation). The legend contains score at a threshold of 20 pixels for each tracker.}
\label{P_C_05}
\end{figure*}
\begin{figure*}[]
\centering
\includegraphics[scale=0.7]{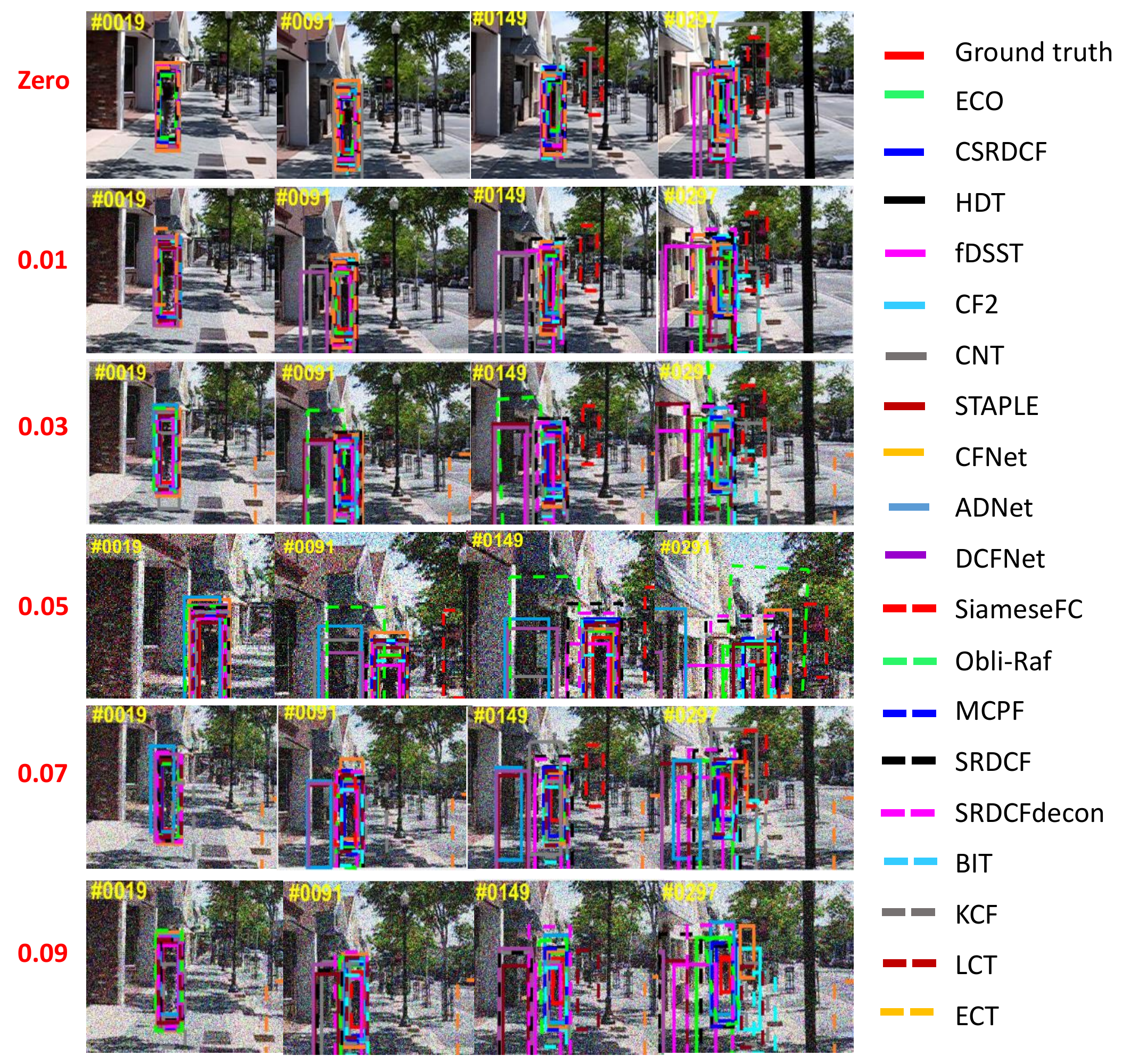}
\caption{Qualitative analysis of CSRDCF \cite{Lukezic_CVPR_2017}, ADNet \cite{yun2017adnet}, CF2\cite{ma2015hierarchical}, DCFNet \cite{wang17dcfnet}, ECO\cite{DanelljanCVPR2017}, CFNet \cite{valmadre2017end}, CNT\cite{zhang2016robust}, KCF \cite{henriques2015high}, HDT\cite{qi2016hedged},  ECT \cite{gao2016enhancement}, Obli-Raf \cite{Zhang2017CVPR}, MCPF \cite{Zhang_2017_CVPR}, SiameseFC \cite{bertinetto2016fully}, SRDCF \cite{danelljan2015learning}, SRDCFdecon \cite{danelljan2016adaptive}, STAPLE\cite{bertinetto2016staple}, fDSST\cite{danelljan2017discriminative}, LCT \cite{ma2015long} and BIT \cite{cai2016bit} over \textit{Human9} sequence containing series of varying noise (left side).}
\label{Human9_N_I}
\end{figure*}
\begin{table*}[t]
\centering
\caption{Speed comparison}
\label{speed_comparision}
\begin{tabular}{|l||c|c|c|c|c|c|c|c|c|c}
\hline
Trackers & CSRDCF   & ECO & HDT  & fDSST & SiameseFC  & CF2 & STAPLE & CFNet & ADNet & \multicolumn{1}{l|}{KCF} \\ \hline
FPS      & 7        & 8   &   5.37   & 96.83 & 25   6.04      &     & 6.51   & 12    & 0.006 & \multicolumn{1}{l|}{65}  \\ \hline
Trackers & Obli-Raf & CNT & MCPF & SRDCF & SRDCFdecon & BIT & DCFNet & LCT   & ECT   &                          \\ \cline{1-10}
FPS      & 0.76     &  0.50   & 0.13 & 1.78  & 1.06       & 30  & 1.5    & 27    & 0.368 &                          \\ \cline{1-10}
\end{tabular}
\end{table*}
\section{Experiments}
In this section, we discuss experimental analysis with  detailed  quantitative and qualitative comparisons. Comprehensive study has been performed on on all the test sequences in object tracking benchmark OTB2015 \cite{wu2015object}, which  consists of 100 sequences,  58,879 frames, and covering eleven different tracking challenges. Six different noisy versions of the OTB2015 are prepared with increasing levels of  additive white Gaussian noise with zero mean and varying variances $\sigma^2=$\{0.00, 0.01, 0.03, 0.05, 0.07, 0.09\}, where 0.00 variance means original dataset. Let  $\mu$ be the mean, and $\sigma^2$ be the variance. Probability of a particular noise vale $x$ is given by
\begin{equation}
P(x| \mu, \sigma) = \frac{1}{{ \sqrt {2\pi \sigma^2 } }}e^{{{ - \left( {x - \mu } \right)^2 } \mathord{\left/ {\vphantom {{ - \left( {x - \mu } \right)^2 } {2\sigma ^2 }}} \right. \kern-\nulldelimiterspace} {2\sigma ^2 }}}.
\end{equation}
For all the compared methods, we have used default  parameters for the experimental investigation, as recommended by the original authors. All experiments are performed on a machine with Intel(R) Core(TM) i5-4670CPU @ 3.40GHz and 8 GB RAM. Execution time comparison in Frames Per Second (FPS)  is shown in table \ref{speed_comparision}.\\
\subsection{Evaluation Methods}
We have adopted for the traditional One Pass Evaluation (OPE) technique to test the robustness of trackers against noise. The OPE evaluation runs trackers only once on a sequence. Precision and success plots have been shown to analyse the performance of trackers. For precision, the Euclidean distance is computed between the estimated centers and ground-truth centers, defined as:
 \begin{equation} \label{eq:2}
\delta_{gp} = \sqrt{(x_g-x_p)^2 + (y_g-y_p)^2},     
\end{equation}
where $(x_g,y_g)$ represents ground truth center location, and $(x_p, y_p)$ is the predicted center location of the target in a frame. During  tracking, a tracker may lose true target position, and estimated position may be random, hence tracking performance can not be measured precisely using average error metric. Instead, the use of a percentage of frames whose estimated locations lies within a provided threshold distance from the ground truth can be a better performance metric. 
\begin{equation} \label{eq:3}
p=\frac{\sum_{n=1}^{N} \chi(\delta_{gp}^n)}{N}*100 ,
\end{equation}
\begin{equation}
\label{eq:4}
\chi(\delta_{gp}^n)= \begin{cases} 
1 \text{~~~if~~~} \delta_{gp}^n \le \delta_{th} \\ 
0  \text{~~~~~~~otherwise}
\end{cases},
\end{equation} where $N$ is the total number of frames.
Legends in the precision plots show that precision corresponding to a threshold of $\delta_{th}=20$ pixels.\\
Precision does not give a clear picture of estimated target size and shape because center location error only measures pixel difference. Therefore, a more robust measure known as success has often been used. For success, an overlap score (OS)  between ground truth bounding box and the estimated bounding box is calculated. Let $r_t$ be the target bounding box and $r_g$ be the ground-truth bounding box. An overlap score is defined as:
\begin{equation} \label{eq:5}
o_s = \frac{|r_t \cap r_g|}{|r_t \cup r_g|},
\end{equation}
where intersection and union of two regions is represented by $\cap$ and $\cup$ respectively, while the number of pixels is denoted by  $| \cdot |$.  The overlap score is used to determine whether a tracking algorithm has successfully tracked a target in the frame. IF $o_s$ score is greater then a threshold, then those frames are referred to as successful frames. Similar to precision, the percentage of overlap score is computed as performance metric: 
 \begin{equation} \label{eq:6}
s =\frac{\sum_{i=1}^N \Gamma({o_s}^i)}{N}*100,
\end{equation}
\begin{equation}\label{eq:7}
\Gamma({o_s^i})= 
\begin{cases}
& 1 \text{~~~if~~~} o_s^i \le t_0 \\
& 0  \text{~~~~~~~otherwise}
\end{cases},
\end{equation}
where $t_0$ is the overlap score threshold, and $N$ is  the total number of frames in the sequence. In the success plot, the threshold value $t_0$ varies between 1 and 0, hence producing varying resultant curves. The success rate threshold $t_0$ value is fixed at 0.5 for evaluation.
\subsection{Quantitative Evaluation}
Figures \ref{P_track} and \ref{S_track} demonstrate the overall precision and success performance of all the trackers with and without additive Gaussian noise  over OTB2015. The percentage for precision degradation is computed as: 
\begin{equation} 
dp_{\sigma} =\frac{(p_0-p_\sigma)}{p_0} \times 100,
\label{eq:8}
\end{equation}
and the percentage of success degradation is computed as
\begin{equation} 
ds_{\sigma} =\frac{(s_0-s_\sigma)}{s_0} \times 100,
\label{eq:9} \end{equation}
where $p_0$, $s_0$ are the precision and success at zero noise and $dp_\sigma$, $ds_\sigma$ indicate the  percentage of precision degradation and success degradation for a tracker at noise level $\sigma$ respectively.\\
From the graphs, we can find that all the trackers got relatively good performance on dataset having zero additive noise compared to noisy dataset. Our investigation shows that CSRDCF, CNT and ECO are much less impacted by noise than the other trackers. As the noise increases, the performance of these trackers degrades linearly with  other trackers.  The CNT does not show much  performance loss in noise, as it constructs filters from target patches. The CSRDCF performs better because, it only updates the target binary mask during the model update, and the tracker does not learn its context from noisy information, while ECO maintains an efficient sample space for noisy trackers, thus performing well in a noisy environment. The ECT was unfortunately  performed better against noise as noise variance increases from 0.01 and 0.09, while the KCF was better then the ECT. The ECT  showed their performance less due to limitations of LDA, while KCF tracks fixed-size objects therefore. Therefore, their performances degrades more at different noise levels. Overall, our investigation indicates that the performance of trackers decreases with the addition of noise.\\
The plot in Figure \ref{P_Per_track}  shows the precision of trackers at threshold of $\delta_{th}=$20. By visual inspection, it can be observed that the performance of all the trackers degrades with an increase of noise. The CNT tracker is much less impacted by noise, as its performance decreases in noise from 51.3\% to 44.4\%, and its curve remains almost straight line. Similarly, for CSRDCF, precision decreases linearly from  89.4 to 79.2 with  increasing noise variance. The performance of the ECO also degrades linearly, while most of the trackers loses show an initial exponential loss in performance, which then become linear beyond a noise variance level of 0.05. Overall, the CSRDCF and the ECO performed well even in noise compared with the other tracking algorithms. The Figure \ref{P_Per_track} b plot shows the  precision degradation with an increase in noise.  This Figure  shows that the performance of the ECT decreases abruptly even with an increase of minor noise, presenting the minimum performance. The KCF is the 2nd minimum performing tracker, and its performance decreases linearly with large slope.\\
Similarly, figure \ref{S_Per_track} represents the success percentage, and the percentage of success degradation. Figure \ref{S_Per_track} a  shows the percentage of success of tracker over varying  noise levels. Overall, the success of the ECO is better compared with  the other trackers even in the presence of noise. The CSRDCF was the second-best tracker. All the trackers showed a drop in their success rate in the presence of noise, but the CNT was much less impacted, as its plot is almost linear, but it still performs more less then other trackers. The success rate for almost all the trackers decreases exponentially as the noise increase, and then decrease becomes linear beyond a noise of $\sigma^2=$0.05 except for the CNT, the CSRDCF and  the ECO. Figure \ref{S_Per_track} b  shows the overall percentage of success degradation of tracker in noise.\\
Figure \ref{P_C_05} shows precision distance plots using the OPE for tracking challenges in the presence of additive white Gaussian noise with 0.05 variance. The ECO performed best for fast motion and out of view challenges, while the CSRDCF performed best for all other challenges. The ECO and The CSRDCF  compete to become the number one tracker under  different challenges. The SRDCF ranked second only once for deformation challenge, and the ECO secured 3rd position in the ranking. The ECT remained last for every challenge, while the CNT secured second-last for the fast motion and motion blur challenges.  Otherwise, KCF was the second-last tracker in the ranking with respect to precision. Similarly, Figure \ref{S_C_05} shows the success plot for eleven different object-tracking challenges with additive Gaussian noise of variance 0.05 and zero mean. Success for the CSRDCF is better then for the other trackers for background clutter, deformation, illumination changes, low resolution, and out of view challenges, while the ECO showed the best success for fast motion, motion blur, in plane rotation, occlusion, out of plane rotation, and scale variation challenges. The KCF and the ECT ranked in second-last and last position respectively for all the challenges. From figure \ref{S_C_05} and \ref{P_C_05} , we notice that the KCF performs less because of its heuristic update strategy and fixed-size target tracking. The ECT performed low in the presence of noise for every challenge because of the limitation of LDA. LDA assumes Gaussian likelihood, and fails if the discriminatory information is not found in the variance of data instead of in the mean. LDA is not suitable for the scenarios where there exists  major object variations. One of the main limitation of LDA is that it sensitive to overfitting.\\
Table \ref{fast_clutter} shows the performance of trackers for fast motion and background clutter object tracking challenges. The ECO performed best, with 87.5 and 91.7 percentage precision compared with  other trackers under zero noise for fast motion and background clutter,  respectively. Performance for all the trackers degrades as the noise, but CSRDCF is a better choice in an environment with more noise,  as  its performance degrades from 85.8 to 76.9 for fast motion and 91.4 to 81.5 for background clutter challenges.\\
The ECO  performances is best for illumination variation and occlusion tracking challenges in clean dataset, while the CSRDCF performs better in noisy environments, as shown in the table \ref{illumination_occlusion}. Performance for the ECO degrades abruptly as  minor noise appears, compared with the CSRDCF, the ADNet and the MCPF for illumination variation. The CSRDCF shows better performance for occlusion in a noisy environment, but the ECO also obtains competitive results.
\subsection{Qualitative Evaluation}
For the qualitative study of tracking algorithms, Figure \ref{Q_I_05} shows tracking results on OTB-100 dataset with additive white Gaussian noise of zero mean and 0.05 variance. Random sequences are selected to cover  all the tracking challenges, including \textit{Basketball, Box, Bolt, Diving, Jogging-1, Human9, Freeman4, Matrix, MountainBike, Skating2-1, Skiing, and Soccer}.  The CSRDCF performed well on almost every selected sequence except for the \textit{Freeman4} sequence. Due to the illumination variation of the target in \textit{Skiing}, the CSRDCF was the only tracker  able to track the target, as all of the others trackers failed. The ECO, the CSRDCF, the DCFNet and the SiameseFC  succeeded in tracking the target in a clean environment but failed in noise, except for  the CSRDCF. Thus, noise has a measurable impact on the performance  of trackers. In the MountainBike sequence, the target moves slowly and has a different colour than the background.  Therefore, all trackers performed good even in the presence of noise.\\
Figure \ref{Human9_N_I} shows the qualitative performance of trackers over \textit{Human9} sequence. In this figure, additive white Gaussian noise increases from top to bottom. The position of the estimated bounding box changes in the frames as noise varies. By visual inspection at frames 91, 149 and 297, we observed an interesting phenomena at all noise levels:  except for the CSRDCF and the ECO, all other tracking algorithms lost the position of their target. This is due to fast motion, motion blur, illumination variation, and scale variation challenges. Our qualitative study indicates that performance of tracking algorithm degrades as the noise level increases.
\begin{table*}[t]
\centering
\caption{Precision performance of trackers for fast motion and background clutter challenges over series of additive Gaussian noise with varying variance}
\label{fast_clutter}
\begin{tabular}{|l||l|l|l|l|l|l||l|l|l|l|l|l|}
\hline
\multicolumn{1}{|c||}{Trackers}& \multicolumn{6}{|c||}{Fast Motion}                                                                                                                                      & \multicolumn{6}{c|}{Background Clutter}                                                                                                                               \\ \cline{2-13} 
\multicolumn{1}{|c||}{}     & \multicolumn{12}{c|}{Noise Variance}                                                                                                                                                                                                                                                                                                          \\ \cline{2-13} 
\multicolumn{1}{|c||}{}                          & \multicolumn{1}{c|}{0.00} & \multicolumn{1}{c|}{0.01} & \multicolumn{1}{c|}{0.03} & \multicolumn{1}{c|}{0.05} & \multicolumn{1}{c|}{0.07} & \multicolumn{1}{c||}{0.09} & \multicolumn{1}{c|}{0.00} & \multicolumn{1}{c|}{0.01} & \multicolumn{1}{c|}{0.03} & \multicolumn{1}{c|}{0.05} & \multicolumn{1}{c|}{0.07} & \multicolumn{1}{c|}{0.09} \\ \hline
ECO     \cite{DanelljanCVPR2017}                                        & 87.5                      & 84.7                      & 82.6                      & 81.2                       & 75.8                       & 70.2                      & 91.7                      & 80.0                      & 77.8                      & 69.0                      & 67.4                      & 64.2                      \\ \hline
CSRDCF   \cite{Lukezic_CVPR_2017}                                        & 85.8                      & 84.4                      & 80.2                      & 78.5                      & 77.4                      & 76.9                      & 91.4                      & 88.6                      & 86.4                      & 85.9                      & 82.1                      & 81.5                      \\ \hline
ADNet \cite{yun2017adnet}                                          & 83.5                      & 74.9                      & 72.9                      & 67.4                      & 61.2                      & 63.0                      & 83.7                      & 72.7                      & 65.4                      & 57.7                      & 53.7                      & 54.7                      \\ \hline
MCPF   \cite{Zhang_2017_CVPR}                                         & 80.5                      & 80.1                      & 71.1                      & 64.9                      & 63.7                      & 60.3                      & 82.0                      & 77.8                      & 65.8                      & 55.0                      & 52.8                      & 46.1                      \\ \hline
SRDCF  \cite{danelljan2015learning}                                          & 79.7                      & 76.5                      & 73.5                      & 72.8                      & 69.4                      & 64.2                      & 86.6                      & 76.6                      & 76.8                      & 64.1                      & 65.3                      & 58.4                      \\ \hline
DCFNet  \cite{wang17dcfnet}                                        & 79.5                      & 73.3                      & 66.5                      & 66.7                      & 65.3                      & 67.0                      & 78.7                      & 77.4                      & 64.5                      & 63.1                      & 57.3                      & 57.4                      \\ \hline
SRDCFdecon \cite{danelljan2016adaptive}                                     & 77.8                      & 77.4                      & 72.5                      & 68.6                      & 56.0                      & 53.8                      & 80.1                      & 71.7                      & 68.4                      & 59.7                      & 50.0                      & 52.2                      \\ \hline
CF2  \cite{ma2015hierarchical}                                           & 77.3                      & 72.7                      & 64.5                      & 59.3                      & 59.7                      & 51.6                      & 75.3                      & 63.7                      & 54.0                      & 48.4                      & 47.9                      & 39.9                      \\ \hline
SiameseFC   \cite{bertinetto2016fully}                                      & 73.9                      & 71.3                      & 68.0                      & 67.1                      & 62.1                      & 60.5                      & 69.8                      & 68.1                      & 61.1                      & 57.5                      & 47.2                      & 53.0                      \\ \hline
HDT  \cite{qi2016hedged}                                            & 73.5                      & 71.0                      & 62.1                      & 61.6                      & 62.5                      & 58.8                      & 70.1                      & 61.7                      & 52.3                      & 47.7                      & 49.6                      & 47.0                      \\ \hline
fDSST \cite{danelljan2017discriminative}                                          & 73.1                      & 69.2                      & 55.8                      & 49.4                      & 48.4                      & 45.5                      & 78.0                      & 65.0                      & 51.6                      & 52.9                      & 50.9                      & 50.6                      \\ \hline
STAPLE \cite{bertinetto2016staple}                                         & 70.8                      & 65.5                      & 57.4                      & 55.0                      & 48.5                      & 44.5                      & 74.9                      & 74.2                      & 66.6                      & 65.4                      & 61.1                      & 60.9                      \\ \hline
LCT   \cite{ma2015long}                                           & 68.0                      & 56.4                      & 56.0                      & 54.1                      & 47.8                      & 45.4                      & 73.4                      & 63.5                      & 60.3                      & 50.4                      & 50.1                      & 43.2                      \\ \hline
ECT  \cite{gao2016enhancement}                                           & 67.6                      & 48.3                      & 1.82                      & 10.0                      & 9.00                      & 4.50                      & 74.4                      & 41.7                      & 5.60                      & 1.70                      & 1.10                      & 0.90                      \\ \hline
BIT  \cite{cai2016bit}                                            & 67.2                      & 57.6                      & 48.7                      & 49.2                      & 42.9                      & 43.4                      & 74.5                      & 60.0                      & 60.9                      & 56.2                      & 52.6                      & 53.3                      \\ \hline
KCF  \cite{henriques2015high}                                           & 63.2                      & 49.2                      & 41.0                      & 35.0                      & 34.8                      & 2.91                      & 71.3                      & 49.5                      & 44.3                      & 44.3                      & 39.8                      & 35.5                      \\ \hline
CFNet   \cite{valmadre2017end}                                        & 63.0                      & 56.7                      & 51.7                      & 41.5                      & 42.7                      & 35.5                      & 71.2                      & 68.1                      & 59.7                      & 51.3                      & 47.9                      & 42.3                      \\ \hline
Obli-RaF  \cite{Zhang2017CVPR}                                      & 61.0                      & 66.1                      & 56.7                      & 60.5                      & 60.6                      & 56.4                      & 74.3                      & 65.5                      & 53.6                      & 50.5                      & 51.4                      & 52.9                      \\ \hline
CNT  \cite{zhang2016robust}                                           & 34.3                      & 34.8                      & 34.2                      & 34.2                      & 30.3                      & 2.71                      & 51.5                      & 51.3                      & 45.2                      & 46.5                      & 47.8                      & 44.3                      \\ \hline
\end{tabular}
\end{table*}
\begin{table*}[t]
\centering
\caption{Precision performance of trackers for illumination variation and occlusion challenges over series of additive Gaussian noise with varying variance}
\label{illumination_occlusion}
\begin{tabular}{|l||l|l|l|l|l|l||l|l|l|l|l|l|}
\hline
\multicolumn{1}{|c||}{Trackers}  & \multicolumn{6}{c||}{Illumination Variation}                                                                                                                           & \multicolumn{6}{c|}{Occlusion}                                                                                                                               \\ \cline{2-13} 
\multicolumn{1}{|c||}{}                          & \multicolumn{12}{c|}{Noise Variance}                                                                                                                                                                                                                                                                                                          \\ \cline{2-13} 
\multicolumn{1}{|c||}{}                          & \multicolumn{1}{c|}{0.00} & \multicolumn{1}{c|}{0.01} & \multicolumn{1}{c|}{0.03} & \multicolumn{1}{c|}{0.05} & \multicolumn{1}{c|}{0.07} & \multicolumn{1}{c||}{0.09} & \multicolumn{1}{c|}{0.00} & \multicolumn{1}{c|}{0.01} & \multicolumn{1}{c|}{0.03} & \multicolumn{1}{c|}{0.05} & \multicolumn{1}{c|}{0.07} & \multicolumn{1}{c|}{0.09} \\ \hline
ECO                                             & 88.8                      & 76.4                      & 77.0                      & 72.0                      & 74.1                      & 64.9                      & 87.2                      & 85.0                      & 80.7                      & 77.2                      & 75.4                      & 73.2                      \\ \hline
CSRDCF                                          & 88.1                      & 84.5                      & 80.5                      & 79.2                      & 77.4                      & 78.2                      & 86.3                      & 84.2                      & 81.2                      & 79.0                      & 77.1                      & 73.3                      \\ \hline
ADNet                                           & 83.6                      & 80.1                      & 70.4                      & 64.3                      & 61.3                      & 62.2                      & 73.2                      & 74.6                      & 65.5                      & 66.4                      & 66.6                      & 60.5                      \\ \hline
MCPF                                            & 84.7                      & 81.4                      & 69.8                      & 68.6                      & 64.0                      & 59.2                      & 82.2                      & 73.7                      & 64.1                      & 56.4                      & 53.5                      & 49.5                      \\ \hline
SRDCF                                           & 79.9                      & 73.0                      & 70.0                      & 66.5                      & 66.6                      & 60.4                      & 81.1                      & 76.7                      & 72.2                      & 69.4                      & 70.0                      & 65.6                      \\ \hline
DCFNet                                          & 77.2                      & 73.0                      & 66.6                      & 61.3                      & 60.3                      & 57.7                      & 78.3                      & 75.6                      & 71.4                      & 651                       & 64.1                      & 62.5                      \\ \hline
SRDCFdecon                                      & 79.3                      & 76.8                      & 68.3                      & 59.3                      & 44.8                      & 46.3                      & 74.7                      & 71.7                      & 60.1                      & 58.5                      & 48.7                      & 43.5                      \\ \hline
CF2                                             & 77.4                      & 72.4                      & 63.5                      & 59.9                      & 58.4                      & 50.7                      & 73.5                      & 62.6                      & 57.4                      & 52.7                      & 48.7                      & 48.8                      \\ \hline
SiameseFC                                       & 73.7                      & 68.3                      & 68.7                      & 63.3                      & 54.9                      & 59.7                      & 69.4                      & 71.5                      & 67.7                      & 65.7                      & 58.6                      & 57.5                      \\ \hline
HDT                                             & 74.2                      & 70.9                      & 64.6                      & 60.3                      & 60.9                      & 56.1                      & 68.1                      & 63.7                      & 53.5                      & 49.3                      & 51.4                      & 53.0                      \\ \hline
fDSST                                           & 72.6                      & 67.8                      & 56.3                      & 48.7                      & 52.7                      & 47.0                      & 62.0                      & 60.2                      & 51.5                      & 45.1                      & 50.2                      & 44.5                      \\ \hline
STAPLE                                          & 76.0                      & 68.1                      & 61.0                      & 61.7                      & 57.6                      & 55.1                      & 71.1                      & 66.2                      & 58.0                      & 58.1                      & 53.9                      & 53.9                      \\ \hline
LCT                                             & 72.7                      & 67.9                      & 62.1                      & 57.2                      & 47.5                      & 47.8                      & 66.7                      & 60.6                      & 55.8                      & 52.5                      & 50.3                      & 46.8                      \\ \hline
ECT                                             & 77.4                      & 53.0                      & 16.9                      & 6.60                      & 5.40                      & 3.70                      & 68.2                      & 39.5                      & 12.5                      & 6.90                      & 6.50                      & 3.80                      \\ \hline
BIT                                             & 71.9                      & 60.0                      & 60.1                      & 53.9                      & 49.5                      & 48.8                      & 71.5                      & 57.2                      & 56.6                      & 54.6                      & 52.8                      & 51.6                      \\ \hline
KCF                                             & 72.1                      & 52.8                      & 43.8                      & 39.7                      & 35.5                      & 32.6                      & 63.2                      & 48.5                      & 42.0                      & 35.1                      & 36.8                      & 29.3                      \\ \hline
CFNet                                           & 71.6                      & 67.9                      & 58.0                      & 51.6                      & 49.5                      & 45.8                      & 62.0                      & 59.1                      & 54.5                      & 48.3                      & 44.5                      & 39.7                      \\ \hline
Obli-RaF                                        & 76.5                      & 66.5                      & 59.8                      & 56.4                      & 52.2                      & 50.5                      & 61.0                      & 61.4                      & 55.7                      & 49.5                      & 49.6                      & 53.9                      \\ \hline
CNT                                             & 49.1                      & 48.4                      & 48.8                      & 47.6                      & 45.0                      & 42.0                      & 50.9                      & 49.8                      & 44.3                      & 45.9                      & 44.6                      & 44.4                      \\ \hline
\end{tabular}
\end{table*}

\section{Conclusion} \label{con}
In our study we have addressed the problems (empirical tresholding, scale invariant features extraction and computational efficiency) regarding LBPs and its variants along with their effectiveness. The objective of this study is to investigate the performance of variants of Local Binary Patterns in encoding texture features in facial images and also with few deep learning based methods. Our study contributes in discussion of key feature analysis in texture extraction. Introduction and analysis of Threshold Local Binary pattern and its variants fully highlight its usefulness in the context of feature extraction. While recently evolved methods for FER like deep learning based methods along with their usefulness and limitations are also being discussed in this article.

\section*{Acknowledgments}
This research was supported by Development project of leading technology for future vehicle of the business of Daegu metropolitan city (No. 20171105).

\bibliographystyle{IEEEtranS}
\bibliography{mybib}

\end{document}